\documentclass{article}
\usepackage{srcltx}
\usepackage[english]{babel}
\usepackage{amsmath}
\usepackage{amssymb}
\usepackage{amsfonts}
\usepackage{latexsym}
\usepackage{textcomp}
\usepackage{appendix}
\usepackage{multirow}
\usepackage{booktabs}
\usepackage{rotating}
\usepackage{afterpage}
\usepackage{pdflscape}
\usepackage{url}
\usepackage{array}
\usepackage{marvosym}
\usepackage{graphics}
\usepackage{graphicx}
\usepackage[numbers]{natbib}
\usepackage[table]{xcolor}% http://ctan.org/pkg/xcolor
\usepackage{epsfig}
\usepackage{float}
\usepackage{authblk}
\newcommand{\N}{{\mathbb N}}
\usepackage[caption=false]{subfig}

\usepackage{color}
\usepackage{hyperref} %para hipervínculos
\newcommand{\R}{{\mathbb R}}

\definecolor{Blue}{rgb}{0,0,0.8}

\def\tc{\textcolor}
\begin{document}

\title{Texture Discrimination via Hilbert Curve Path Based Information Quantifiers}

\author[1]{Aurelio F. Bariviera \thanks{Corresponding author: aurelio.fernandez@urv.cat}}
\author[2]{Roberta Hansen}
\author[2,3]{Ver\'onica E. Pastor}

\affil[1]{\scriptsize  Universitat Rovira i Virgili, Department of Business, ECO-SOS, Av. Universitat 1, 43204 Reus, Spain.}
\affil[2]{\scriptsize  Universidad de Buenos Aires, Facultad de Ingenier\'ia, Departamento de Matem\'aticas, Argentina.}
\affil[3]{\scriptsize Universidad Nacional de La Plata, Facultad de Ingenier\'ia,\ Departamento de Ciencias B\'asicas, Argentina.}
\maketitle

\begin{abstract}
The analysis of the spatial arrangement of colors and roughness/smoothness of figures is relevant due to its wide range of applications. This paper proposes a texture classification method that extracts data from images using the Hilbert curve. Three information theory quantifiers are then computed: permutation entropy, permutation complexity, and Fisher information measure. The proposal exhibits some important properties: (i) it allows to discriminate figures according to varying degrees of correlations (as measured by the Hurst exponent), (ii) it is invariant to rotation and symmetry transformations,  (iii) it can be used either in black and white or color images. Validations have been made not only using synthetic images but also using the well-known Brodatz image database. 


\end{abstract}

%%%%%%%%%%%%%%%%%%%%%%%%%%%%%%%%%%%%%%%%%%%%%%%%%%%%%%%%%%
\section{Introduction}
%%%%%%%%%%%%%%%%%%%%%%%%%%%%%%%%%%%%%%%%%%%%%%%%%%%%%%%%%%%
Texture is a fundamental property of images that describes the spatial arrangement of intensities or colors. It plays a crucial role in various scientific fields, including object recognition, medical image analysis, and product quality control. Accurate texture classification is essential for tasks such as identifying objects in images, detecting abnormalities in medical scans, and ensuring the consistency of manufactured products. However, traditional methods for extracting and analyzing textures can be computationally expensive and susceptible to image quality issues such as noise, blur, and color variations.

To overcome these limitations, there is a growing need for robust and efficient texture classification techniques. These methods should be able to accurately discriminate textures from various sources, including natural product images such as those of rocks, fabrics, and skin, which exhibit complex and diverse textures. A successful texture classification method should be able to effectively distinguish between them, even under varying image quality conditions. Moreover, in image classification, rotational invariance is a desirable feature in algorithms. This means that the algorithm can accurately classify images regardless of their orientation or rotation \cite{Hamou-Aouat-2017}. Thus, a reliable method should account both for image rotation and other types of rigid transformations.

 The development of such methods has the potential to revolutionize various scientific domains. In object recognition, texture classification can aid in identifying objects based on their unique textural signatures. In medical image analysis, it can help in detecting abnormalities and assessing tissue health by analyzing the textures of diseased tissues. In product quality control, texture classification can be used to ensure consistent product quality by identifying deviations in texture patterns. In the fashion industry, texture classification could be a valuable tool for identifying and understanding consumer preferences for different types of garments. 
 
 The use of information-theory-related methods in one dimension is not new in the analysis of time series. In fact, it has been successfully used for the analysis of dynamical systems in different scientific domains. For example, 
\cite{Rosso2002} use wavelet entropy to characterize electroencephalography records across different stages of brain activities; \cite{DeMicco2008} propose the use of permutation entropy and statistical complexity for selecting the most convenient randomizing technique; in a similar vein, \cite{Saco2010} explores the relationship between entropy and epochs of rapid climate change in the Holoce \cite{Rosso2013} highlights the robustness of permutation entropy and statistical complexity quantifiers to characterize a long list of nonlinear chaotic maps. \cite{Soriano2011} find that permutation entropy and complexity reveal key dynamical features of a semiconductor laser.
Finally, \cite{Bariviera2015} unveil interest rate manipulation using global and local information theory quantifiers. 

 However, as \cite{Feldman2003} recall, there is a need to study spatial structure and pattern in more than one dimension. In particular, it is not trivial to understand how information is assembled across a two-dimensional lattice, and how to extract such information. By ``global'' and ``local'' quantifiers we mainly refer to the Shannon entropy and the Fisher information measure, respectively. 
 Motivated by this fact, Ribeiro and coauthors  \cite{Ribeiro2012}, introduced an extension of the complexity-entropy causality plane to evaluate the complexity of two-dimensional patterns, as are fractal images, liquid crystal textures and Ising surfaces. Roughly speaking, they consider an image as a $m\times n$ matrix that is split into $D_m\! \times\! D_n$ submatrices. Each submatrix is converted, following a row-wise order, into a symbol and then, they derive a time series from the calculated symbols and compute the permutation information theory quantifiers. Finally, they
 use the complexity-entropy causality plane to classify images. Later, Zunino and Ribeiro  \cite{ZUNINO2016679}, applied this extension to study texture classification in real images. Sigaki et al. \cite{Sigaki2018} also applied this extended method to analyze different styles in historic art paintings and classify their artworks. It is important to mention that these works approach the topic of higher dimensions only for the complexity quantifier, which is interpreted, as said before, as a global one.  
 Up to our knowledge, there is no proposed method extended to higher dimension arrangements to calculate the Fisher Information Measure, neither the corresponding Fisher-entropy plane.  
 The value of the Fisher Information Measure defined in Eq. \ref{Fisher-disc}, for a discrete probability distribution $P\!=\!\{p_i\}_{1\le i\le N}$, strongly depends on the $i$-ordering --hence its condition as a local quantifier. The key point is that there is no a uniquely predetermined order, when $P$ is defined on two --or higher-- dimension patterns, to convert a matrix into a symbol.

 In this sense, this paper presents a novel two-step method that first transforms a two-dimensional image into a one-dimensional time series using the Hilbert curve and then computes three information theory quantifiers using the Bandt \& Pompe symbolization method. The method extracts the intrinsic characteristics of the images and allows their classification and discrimination. The Hilbert curve was chosen to effectively transform spatial correlations into temporal ones.
 %The election of the Hilbert curve arises from the need to convert, as best as possible, spatial correlations into temporal ones.

 We first evaluate the proposed method on images generated by self-similar multifractal surfaces and by Brownian surfaces, demonstrating, in the last case, its ability to accurately classify textures based on their Hurst exponents. Next, we apply the method to the Brodatz image database, which has been widely used in texture classification studies, and demonstrate its robustness to image rotation, mirror-symmetry, and color variations. %% referencia de Roberta

Our contribution is based on the development of a novel method for analyzing images by transforming them into a regular time series that offers several advantages over existing approaches: (i) It eliminates directional biases inherent in row/column-wase scan, ensuring a more balanced representation of the image's texture. (ii) It leverages both the Hilbert curve's space-filling and non-privileged direction properties to generate a path that visits each pixel once, ensuring no data redundancy. (iii) It utilizes information theory quantifiers, permutation entropy, permutation statistical complexity and Fisher information measure, to provide a robust and informative classification of texture classes. (iv) it results robust under rigid images transformations. (v) it can be generalized in a natural way to a more than two dimensions and be effectively applied to classify $n$-dimensional patterns.
 
 The remainder of the paper is organized as follows. Section \ref{sec:methods} provides a summary of the information  quantifiers considered, followed by Sec. \ref{sec:proposal} where the proposed method is presented. Section \ref{sec:results} discusses the key findings and finally Sec. \ref{sec:conclusions} summarizes the paper's contributions.

%%%%%%%%%%%%%%%%%%%%%%%%%%%%%%%%%%%%%%%%%%%%%%%%%%
\section{Statistical Information Quantifiers \label{sec:methods}}
%%%%%%%%%%%%%%%%%%%%%%%%%%%%%%%%%%%%%%%%%%%%%%%%%%

This section describes the information theory quantifiers that will be used in the empirical section of the paper. 

\subsection{Shannon Entropy}
In the context of discrete random variables, we consider an arbitrary variable denoted as $X$. Our primary objective is to quantify the information content obtained upon observing this variable. The quantification of this information relies upon the associated probability distribution, denoted as $p(x)$. In the realm of dynamical systems, Shannon entropy emerges as a metric capturing the overarching smoothness of the probability distribution $p(x)$, exhibiting invariance to alterations in the arrangement of probability density values concerning the independent variable. It inherently characterizes the degree of disorder within the system, with respect to the likelihood of various accessible states. In situations where an event is certain ($p(x)\!=\!1$), the information gained is minimal, as the certainty of its occurrence was already established. Conversely, in the case of improbable events, the revelation of such occurrences imparts a substantial amount of information. Indeed, the highest level of uncertainty regarding an event manifests itself when the probability distribution is uniform. This conceptual framework, inspired by Shannon's seminal work \cite{book:shannon1949}, leads to the formulation of Shannon's Entropy as follows:

\begin{equation}
S[P]=-\sum_{j=1}^N p_j \ln(p_j)
\label{eq:Shannon}
\end{equation}

Here, $P\!=\!p_j$, $j\!=\!1,\dots, N$ represents a specific probability distribution.

Shannon's Entropy serves as a ``global'' measure and exhibits limited sensitivity to localized perturbations within the distribution. To mitigate this limitation and facilitate comparative analysis, Shannon's Entropy can be normalized to a range of values between 0 and 1. This normalization is achieved by dividing the entropy by its maximum attainable value, defined as:

\begin{equation}
{\cal H}[P]=\frac{S[P]}{S_{\max}}=\frac{-\sum_{j=1}^N p_j \ln(p_j)}{\ln N}
\label{eq:ShannonNormalized}
\end{equation}

In this normalized form, denoted as ${\cal H}[P]$, the entropy's values are rescaled to a dimensionless range, facilitating the examination of information content across different probability distributions.

\subsection{Statistical complexity}

Examining time series solely through Shannon entropy measures may not provide a comprehensive understanding of their underlying dynamics. As \cite{FeldmanCrutchfield98} point out, entropy alone fails to capture the intricate organization and structure that often characterize complex systems. To gain a more holistic perspective, statistical complexity measures come into play. These measures can uncover hidden patterns and correlations within time series, allowing for a better characterization of the system's behavior. Statistical complexity measures assign higher values to time series that exhibit a rich interplay of patterns, reflecting a greater degree of structure and organization. Conversely, highly ordered or random sequences, such as perfectly periodic or completely unpredictable patterns, are attributed lower complexity values. This distinction highlights the ability of statistical complexity measures to distinguish between systems that exhibit varying levels of internal organization.

Based on the functional form defined by \cite{LMC95}, \cite{Lamberti2004} introduced a complexity measure that reads:
\begin{equation}
 {\cal C}_{JS}[P]={\cal Q}_J[P,P_e] \,{\cal H}[P]   
\end{equation}
where $P_e$ is the uniform distribution, and ${\cal Q}_J$ is the disequilibrium between the observed and the uniform probability distributions, defined in terms of the Jensen-Shannon divergence.

\subsection{Fisher Information Measure}

In a landmark contribution to statistical science, \cite{Fisher1922} introduced the concept that nowadays is known as {\it Fisher Information Measure} (FIM), a versatile metric that quantifies the amount of information contained within a set of data about an unknown parameter. Beyond its role in parameter estimation, FIM serves as a measure of the informational yield of a data set, reflecting the `quality' of the measurements and providing insights into the underlying system or phenomenon \citep{Frieden1998}.

Unlike Shannon entropy, FIM distinguishes itself by incorporating derivative terms, effectively gauging the sensitivity of the probability density function to reordering over the independent variable. This heightened sensitivity proves particularly valuable in scenarios where a clear ordering exists, such as in dynamic systems where time naturally serves as an ordering variable. Thus, FIM quantifies the ``gradient content'' of a distribution and reads:
\begin{equation}
    {\cal F}[f]=\int_{\Delta} \frac{1}{f(x)}\left[ \frac{df(x)}{dx}\right]^2 dx = 4 \int_{\Delta} \left[\frac{d \psi(x)}{dx} \right]^2
\end{equation}

The standard FIM definition involves dividing by the distribution itself, which becomes problematic when the distribution approaches zero. Using real probability amplitudes ($\psi(x)$) avoids this issue and shows FIM simply measures the gradient in $\psi(x)$ \citep{Fisher1922,Frieden1998}. This sensitivity to local variations makes FIM a ``local'' quantifier.

For discrete systems with $N$ states, discretizing FIM introduces complications like losing shift-invariance. We focus on a normalized discrete FIM $(0 \leq {\mathcal F}\leq 1)$ given by:
\begin{equation}
\label{Fisher-disc}
{\mathcal F}[P]~=~F_0~\sum_{i=1}^{N-1}\big[(p_{i+1})^{1/2} - (p_{i})^{1/2}\big]^2 
\end{equation}
This discretization is considered well-behaved in discrete settings \citep{Dehesa2009}. The normalization constant $F_0$ reads:
\begin{equation}
\label{F0}
F_0~=~\left\{
       \begin{array}{cl}
                    1       &\quad \mbox{if $\,p_{i^*}\!=\! 1$, for
                            $i^*\!=\! 1$ or $i^*\!=\! N$ and $\,p_{i}\!=\!0 \ \ \forall\,  i\!\neq\! i^*$} \\
                    1/2     &\quad \mbox{otherwise}
       \end{array}
\right.
\end{equation}

A highly ordered system (most $p_i$ near zero, one $p_k$ near 1) has low Shannon entropy (${\mathcal H}\!\sim \!0$) and high normalized FIM (${\mathcal F}\!\sim\!1$). Conversely, a disordered system (all $p_i$ similar) has high entropy (${\mathcal H}\!\sim \!1$) and low FIM (${\mathcal F}\!\sim \!0$). This discrete FIM behaves opposite to Shannon entropy, except for periodic motions \citep{OPR2012,Olivares2012B}. 

The addings capture a notion of roughness in the probability density function, quantified by the difference between adjacent probability states. This necessitates defining an order in which these states appear. Related literature often relies on either the order proposed by \cite{Keller2005} or the ``Lehmer code" described by \citep{Lehmer}. The latter employs a lexicographic ordering generated by an algorithm based on the factoradic number system. This paper follows this ordering, based on previous results in the literature \citep{OPR2012,Olivares2012B,Bariviera2015}.

%%%------------------------------------------------
\subsection{Bandt and Pompe symbolization method}

It is imperative to establish the probability density distribution of the system under investigation to calculate the aforementioned quantifiers. There are several alternative procedures to complete this task, such as amplitude-based procedures \citep{DeMicco2008}, binary symbolic dynamics \citep{Mischaikow}, among others.
However, two decades ago, Band \& Pompe (BP) \cite{BandtPompe02} introduced a powerful technique that has found extensive applications in various fields, including physics, biology, finance, and climate science. This method, based on the ordinal relations between neighboring values, transforms the observed time series data into symbolic sequences. Such transformation enables the extraction of valuable features and patterns from complex temporal data.

The BP technique operates as follows. Let consider a time series ${\cal S}(t)\!=\!\{x_t\}_{1\le t \le N}$, an embedding dimension $D\!>\!1$, and an embedding delay $\tau$, with $D,\tau\!\in\!\N$. Then, we are in a position to define patterns of order $D$, generated by:
\begin{equation}
    {\cal s} \mapsto \big(x_{s-(D-1)\tau},x_{s-(D-2)\tau},\dots,x_{s-\tau},x_s\big)
\end{equation}
Then, patterns are evaluated according to the ordinal position of their components. For example, let $\{1,6,3,2,8,9\}$ be a time series, $D\!=\!3$ and $\tau\!=\!1$ the embedding and delay dimensions. The following patterns are constructed: $(1,6,3)$, $(6,3,2)$, $(3,2,8)$, $(2,8,9)$, which are transformed into the following symbols  by substituting the original values by the respective rankings: $(0,2,1)$, $(2,1,0)$, $(1,0,2)$, $(0,1,2)$. 
Therefore, BP transforms the original time series into a series of symbols of length $D$. These symbols, by the nature of their construction, include temporal information, which grows along with $D$.  

Given an embedding dimension $D$, there are $D!$ patterns. The frequency of appearance of these patterns in the time series could be used as a straightforward measure to construct the probability density function associated to the symbolic time series $P\!=\!\big\{p(\pi_i),\,i\!=\!1,\dots, D!\big\}$, where:
\begin{equation}
   p(\pi_i)=\frac{\#\big\{s\!:\,s\leq N\!-\!(D\!-\!1)\,\tau;\ (s) \mbox{ has type } \pi_i\big\}}{N\!-\!(D\!-\!1)\,\tau} 
\end{equation}

It was shown in previous research that stochastic processes include all permutation patterns in their respective PDF. In other words, there are no ``forbidden patterns'', and the relative frequency of each pattern will depend on the type of stochastic process under analysis. On the contrary, chaotic processes could include them, as some orbits are never observed due to their deterministic nature. 

The BP approach has several advantages over conventional methods based on range partitioning. It is invariant concerning nonlinear monotonous transformations, meaning that nonlinear drifts or scalings artificially introduced by a measurement device will not modify the quantifiers' estimation. This is a desirable property if one deals with experimental data. The BP method is also simple and fast, requiring only two parameters: the pattern length/embedding dimension $D$ and the embedding delay $\tau$. Finally, the BP method can be applied to any type of time series, including regular, chaotic, noisy, or reality-based. The only condition for the applicability of the BP method is a very weak stationary assumption, which states that the probability for $x_t\!<\! x_{t+k}$ should not depend on $t$ for $k\!=\!D$.
  
Overall, the BP method of symbolic analysis of time series has proven to be a versatile and robust tool for characterizing complex signals. It offers a unique perspective on the underlying dynamical system, gaining deeper insights into a wide range of phenomena. 

%%%---------------------------------------------------
\subsection{Complexity Entropy and Entropy Fisher Causality Planes}

In a breakthrough paper, Rosso and coworkers \cite{RossoNoise07} showed that the joint study of the permutation entropy and permutation statistical complexity provides useful information to discriminate stochastic and chaotic processes. 
Causality indicates the temporal correlations between successive samples in the ordering suggested by Bandt and Pompe.

For each probability distribution, $P\!=\!\{p(\pi)\}$,  the pair $\big({\cal H}[P],{\cal C}_{JS}[P]\big)$ is represented in the plane ${\cal H}\!\times\! {\cal C}$, so called {\it Complexity Entropy Causality Plane}  (CECP). Although such pairs must belong to $[0,1]\!\times\![0,1]$, due to their normalized components, 
Martin, Plastino and Rosso \cite{Martin2003} explained that in the CECP, all the possible values are bounded by two continuous curves representing the maximum and minimum values of ${\cal C}_{JS}$ as a function of $\cal H$. The CECP has been effectively used to rank stock markets according to their informational efficiency \cite{ZUNINO2010efficiency}, to classify commodities \cite{FERNANDES2020109909}, to unveil the chaotic dynamics of a semiconductor laser \cite{Soriano2011}. 

An additional insight into the hidden characteristics of dynamical systems could be provided by the so called {\it Fisher Entropy Causality Plane} (FECP), introduced by Vignat and Bercher \cite{Vignat2003}, in which the pair $\big({\cal H}[P],{\cal F}[P]\big)$ is represented in the plane ${\cal H}\!\times\!{\cal F}$. It was used, for example, to detect the change in the observed stochastic process due to manipulation of data \cite{Bariviera2015}. 

Given the relevance of both global and local perspectives, in the present article we will classify textures using both planar representations.

%%%%%%%%%%%%%%%%%%%%%%%%%%%%%%%%%%%%%%%%%%%%
\section{Proposed method \label{sec:proposal}}

It seems reasonable to assume that the CECP can be used to analyze properties of images, considering the value and relative arrangement of their pixels. The key problem is how to extend the ordering proposed by Bandt \& Pompe from 1D to 2D, or conversely, how to transform a 2D image into a 1D time series. 
Transforming a digitized surface into a 2D image necessitates a strategic reading approach to extract meaningful information. Conventional methods, such as reading by rows or simultaneously processing multiple rows, overlook the inherent spatial relationships between pixels, introducing potential biases due to preferential reading directions.

 To address these limitations, we propose a novel methodology that traverses the entire image in a non-repeating, direction-diverse manner, ensuring that each pixel is visited once and only once. We achieve this by employing the Hilbert curve, a space-filling curve that generates a path that visits every pixel while maintaining a high degree of spatial locality.

%%%---------------------------------------------
 \subsection{The Hilbert curve} 
\begin{figure}[!ht]
\hspace{-0.5cm}
\includegraphics[scale=0.35]{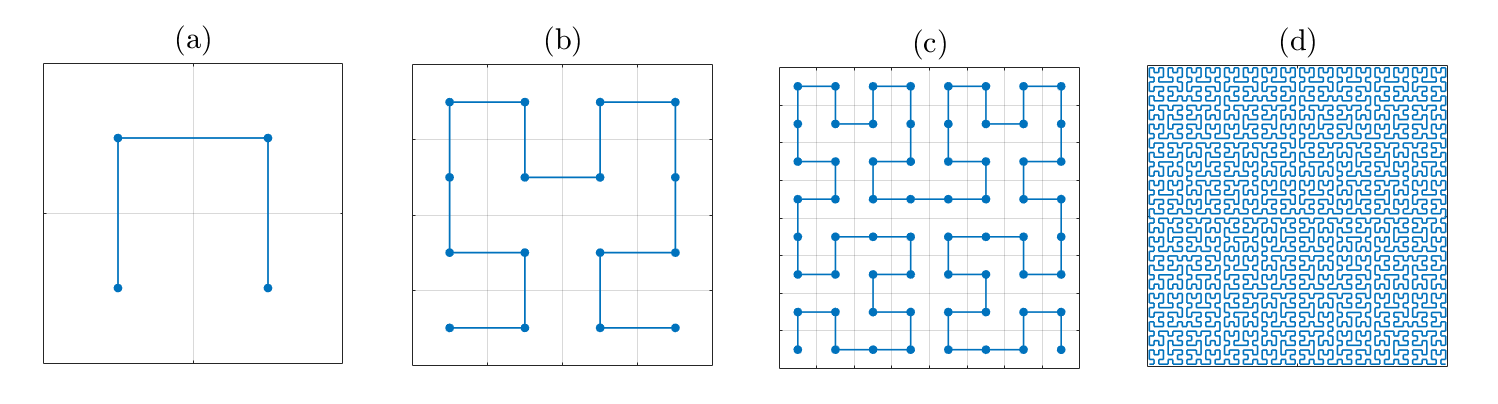}
\caption{Levels of planar Hilbert curve. (a) level 1, (b) level 2, (c) level 3, (d) level 6.}
\label{hilbert-levels}
\end{figure}

This well-known curve, is a fractal continuous curve in $\mathbb{R}^n$, with full Hausdorff dimension $n$, due to its space-filling property. The curve is obtained as a limit of polygonal curves of increasing lengths, called {\it levels}, as it is shown in Fig. \ref{hilbert-levels} for $n\!=\!2$. Notice that the higher the level the more space is filled by the polygonal curve. If the unit square is divided into $2^N\!\!\times\!2^N$ sub-squares, then a Hilbert curve of level $N$ is needed to scan all sub-squares.

A very important feature of space-filling curves is that they can map multidimensional data to one dimension while preserving locality of the data points \cite{Moon}. This means that two close points in a one dimensional arrangement correspond to two close points in the $n$-dimensional arrangement before unfolding (the converse is not true). This can be seen in Fig. \ref{hilbert-levels} by noting that consecutive nodes in a polygonal curve, correspond to neighboring sub-squares. 
%%%%----------------------------------
%%%---------------------------------------
This property is very convenient when the local correlation of the data is to be preserved. Additionally, while other curves can also unfold a two-dimensional arrangement with this property, the Hilbert curve does  so more effectively. This efficiency comes from its self-similarity, which follows a recursive pattern, and it accomplishes this fact without favoring any particular direction of travel.
These features make the Hilbert curve to be widely used in data based applications, as multi-dimensional indexing methods, data structures, parallel computing, image processing and encryption algorithm \cite{Dai,Tang}.

 The resulting time series, extracted from the Hilbert curve-based scan, captures the spatial characteristics of the image while mitigating the inherent biases of conventional methods. We then employ information theory quantifiers described in Section \ref{sec:methods}, using the CECP and FECP, to classify the time series into distinct texture classes. These planar representations provide a comprehensive and insightful understanding of the texture characteristics.

%%%=============================================================
\section{Results \label{sec:results}}

The proposed method is tested on simulated images and in a widely used texture database, namely: (i) self-similar multifractal surfaces, (ii) Brownian surfaces, (iii) Black and white Brodatz dataset, and (iv) colored Brodatz dataset. The experiments not only include varying the value of the parameters in the calculation of the information quantifies, but also include the rotation and mirror-symmetry transformation of the images. 

%%%%===========================================================
\subsection{Self-similar Multifractal Measures as Complex Images}

The complexity of an image is determined by the spatial distribution of a scalar quantity, as it could be the color or gray-level at every pixel. This quantity, when normalized, can be considered as a probability mass distribution, $\mu$, supported on the image. The way $\mu$ is spread on the region characterizes the type of image in question. When the mass density is so highly irregular that its concentration obeys different local power laws, and at the same time, the sets of points having the same local mass concentration define different fractals subsets on the support, then $\mu$ is called a {\it multifractal measure}. It posses a very rich structure at every scale which yields to a whole set of fractal dimensions, usually studied by means of various classes of multifractal spectra \cite{Falconer}. 

In this section we investigate a  multifractal mass distribution defined on the unit square as a planar image, by means of information-theory methods rather than by terms of a classical multifractal dimensional analysis. 

\subsubsection{Multinomial multiplicative cascades}
An example of a process generating a self-similar multifractal measure (SMM), is a deterministic one, where an euclidean set, $E\!\subset\!\mathbb{R}^n$, is fragmented into smaller equal-size pieces according to a recursive fixed rule, and simultaneously, the total mass ($\mu(E)\!=\!1$) is distributed on the pieces following another non trivial recursive rule. This process is called a {\it multiplicative cascade}, and generate a SMM called a {\it multinomial measure}, which has features typical of a large class of multifractal ones \cite{Evertsz-Mandel}.

\begin{figure}[ht]
\begin{center}
\includegraphics[scale=0.45]{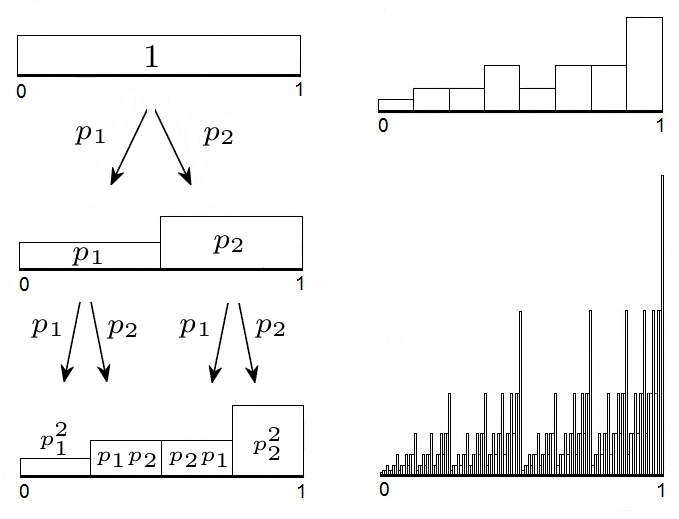}
\end{center}
\caption{Construction of a SMM $\mu\!=\![p_1,p_2]$ (binomial cascade) on the unit interval with $p_1+p_2\!=\!1$, $p_i\!>\!0,i\!=\!1,2$, starting from the uniform distribution. Scheme of the first three steps  and an advanced step of the recursive process are shown (see the text for the explanation).}
\label{binomial}
\end{figure}

To illustrate the process, let $E\!=\![0,1]\!\subset\!\R$ and $p_1,p_2\!>\!0$, with $p_1+p_2\!=\!1$. A  binomial SMM, $\mu\!=\![p_1,p_2]$, is constructed on $E$, as follows: at first step, $E$ is divided into two sub-intervals of length $1/2$, and an initial uniform density is redistributed by assign mass $p_1$ to the left sub-interval, and mass $p_2$ to the right one. In the following steps, the equal-size subdivision of intervals are applied successively, the same as the mass on each sub-interval is rescaled by a factor of $p_i$, as it is, left $i\!=\!1$, or, right $i\!=\!2$. The process is illustrated in Fig. \ref{binomial}. The last panel exhibits the extreme irregularity of measure $\mu$.
%%%%-------------------------------------------
%%%-----------------------------------------------------------------
\subsubsection{Worked example}

\begin{figure}[ht!]
\hspace{-0.5cm}
\includegraphics[scale=0.33]{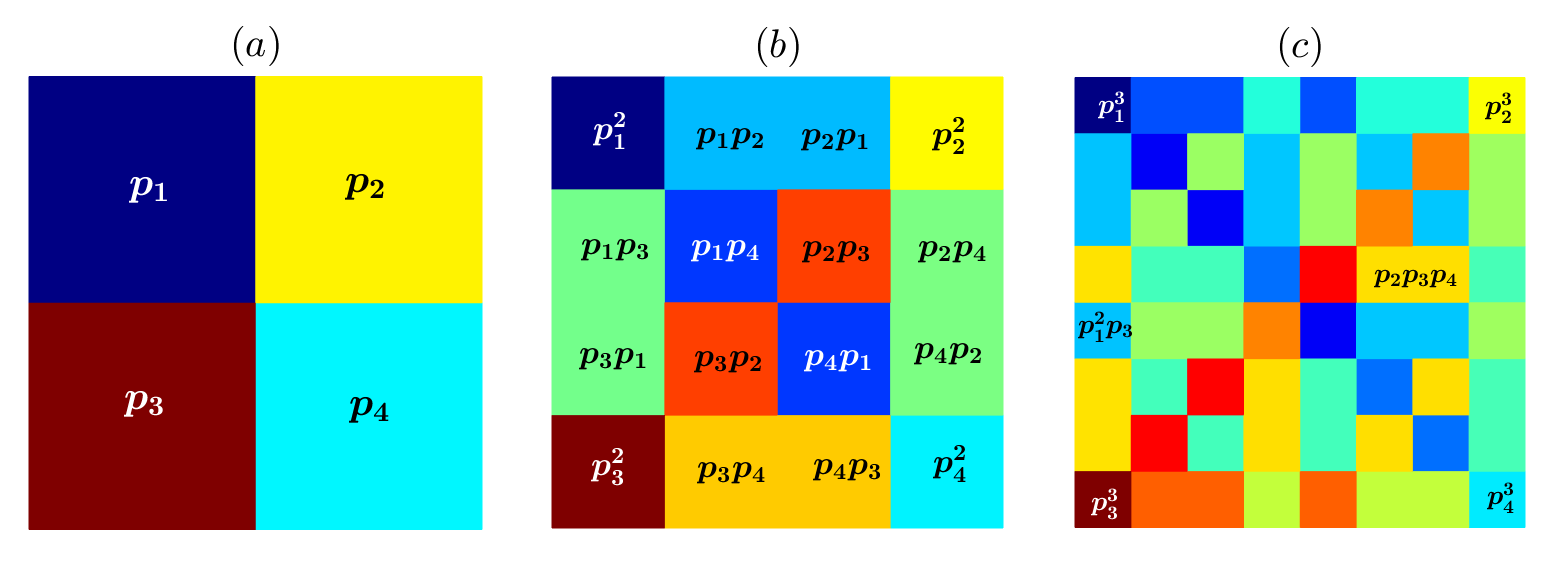}
\caption{Three steps of SMM construction, $\mu\!=\![p_1,p_2,p_3,p_4]$ on the unit square, $p_1\!=\!0.2434,p_2\!=\!0.2522,p_3\!=\!0.2566$ and $p_4\!=\!0.2478$. (a) Measure values attached to sub-squares at step 1, (b) The same but at step 2, (c) The same but at step 3 (only a few). The colours are in accordance to the measure values.}\label{cuatrinom}
\end{figure}

\begin{figure}[ht!]
\includegraphics[scale=0.32]{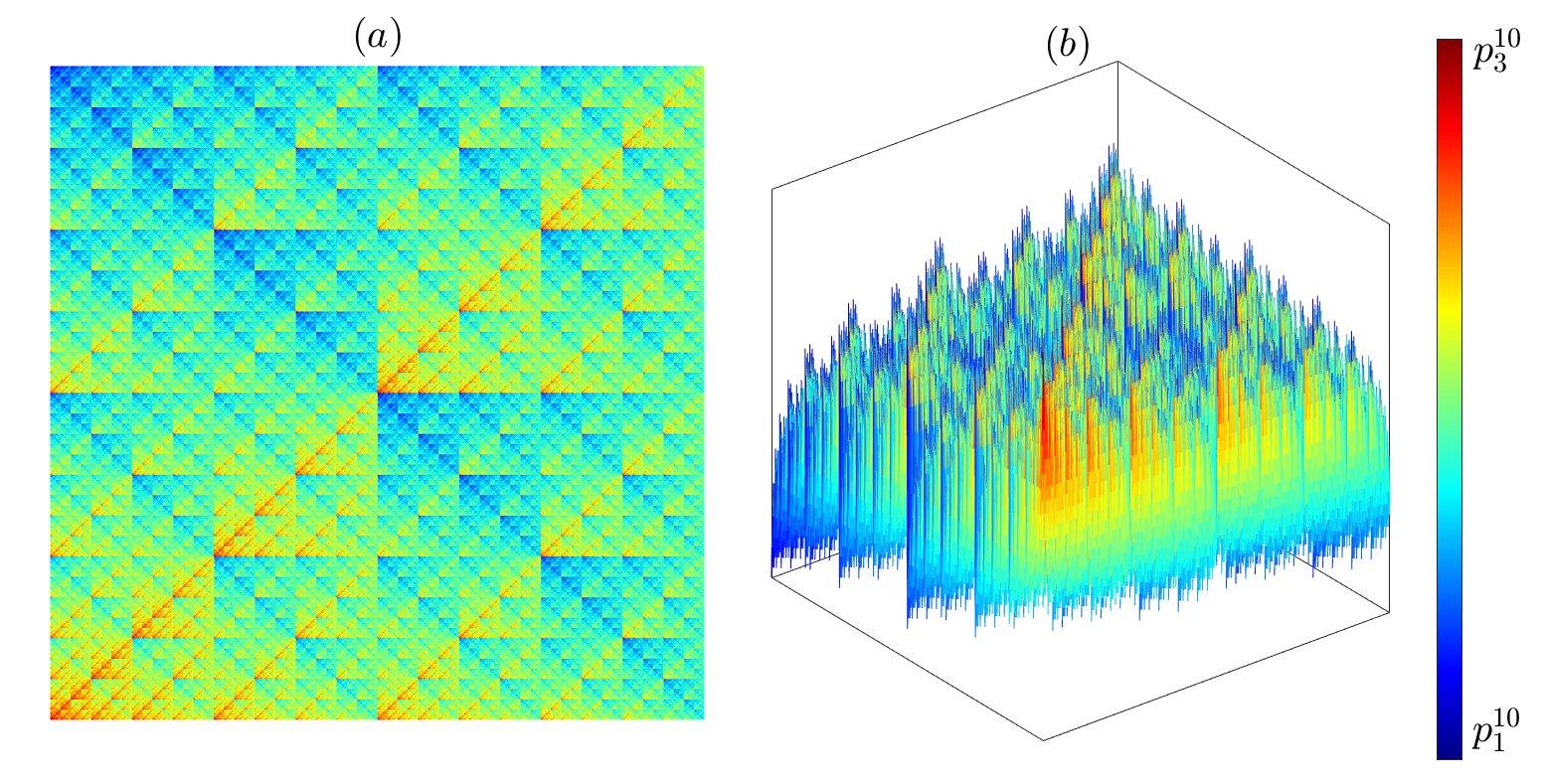}
\caption{Multifractal-type image. (a) Measure values attached to sub-squares at step 10 (only the colours), (b) 3-dimensional rendering of (a) showing the SMM $\mu$. The colours are in accordance to the measure values. ($p_1\!<\!p_4\!<\!p_2\!<\!p_3$).}
\label{cuatrinom-2}
\end{figure}

The analogous of the binomial SMM constructed on $[0,1]$, for the two dimensional case, is a similar construction but on the unit square, $E\!=\![0,1]\!\times\![0,1]\!\subset\!\R^2$. We will consider a quatrinomial SMM, $\mu\!=\![p_1,p_2,p_3,p_4]$, obtained from the same recursive process described above, by dividing each square into four equal size sub-squares, with the values: $p_1\!=\!0.2434,p_2\!=\!0.2522,p_3\!=\!0.2566$ and $p_4\!=\!0.2478$ ($\sum_{i=1}^4p_i\!=\!1$) (see Fig. \ref{cuatrinom}). The image obtained at the $10^{\rm th}$ step of the process, and shown in Fig. \ref{cuatrinom-2}(a), will be used as example of a multifractal-type image.
It consists of a $2^{10}\!\times\!2^{10}$ arrange of  $\mu$-values which will be scanned by a Hilbert curve of level 10 and then unfolded to get the corresponding time series.

%%%--------------------------------------------
%%%-------------------------------------------------
%%%------------------------------------------------

\begin{figure}[ht!]
\includegraphics[scale=0.32]{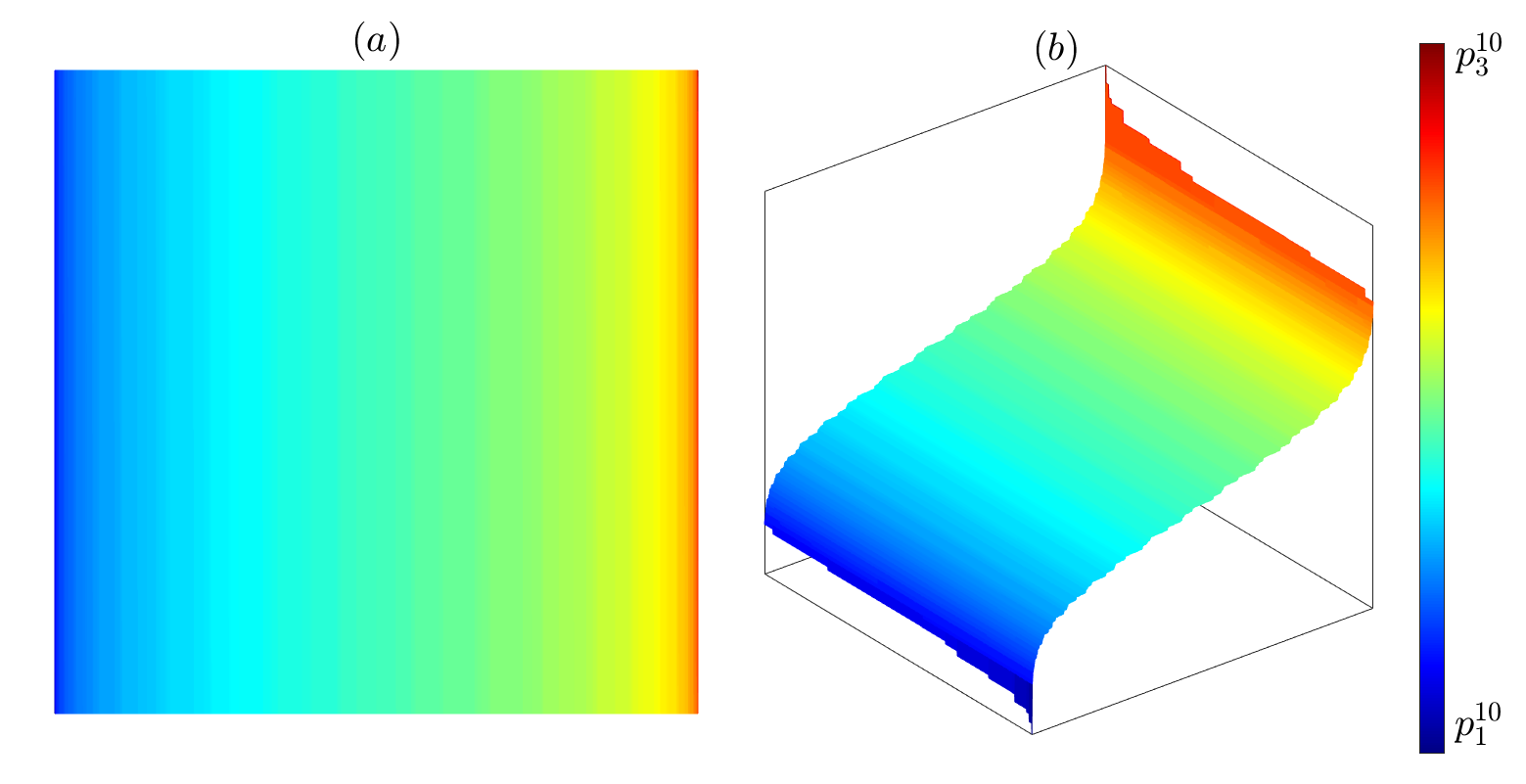}
\caption{(a) Regular image obtained by ordering the $\mu$-values of the multifractal surface in Fig. \ref{cuatrinom-2} from lowest to highest, (b) 3-dimensional rendering of (a).}
\label{cuatrinomial-ordered}
\end{figure}

\begin{figure}[ht!]
\includegraphics[scale=0.32]{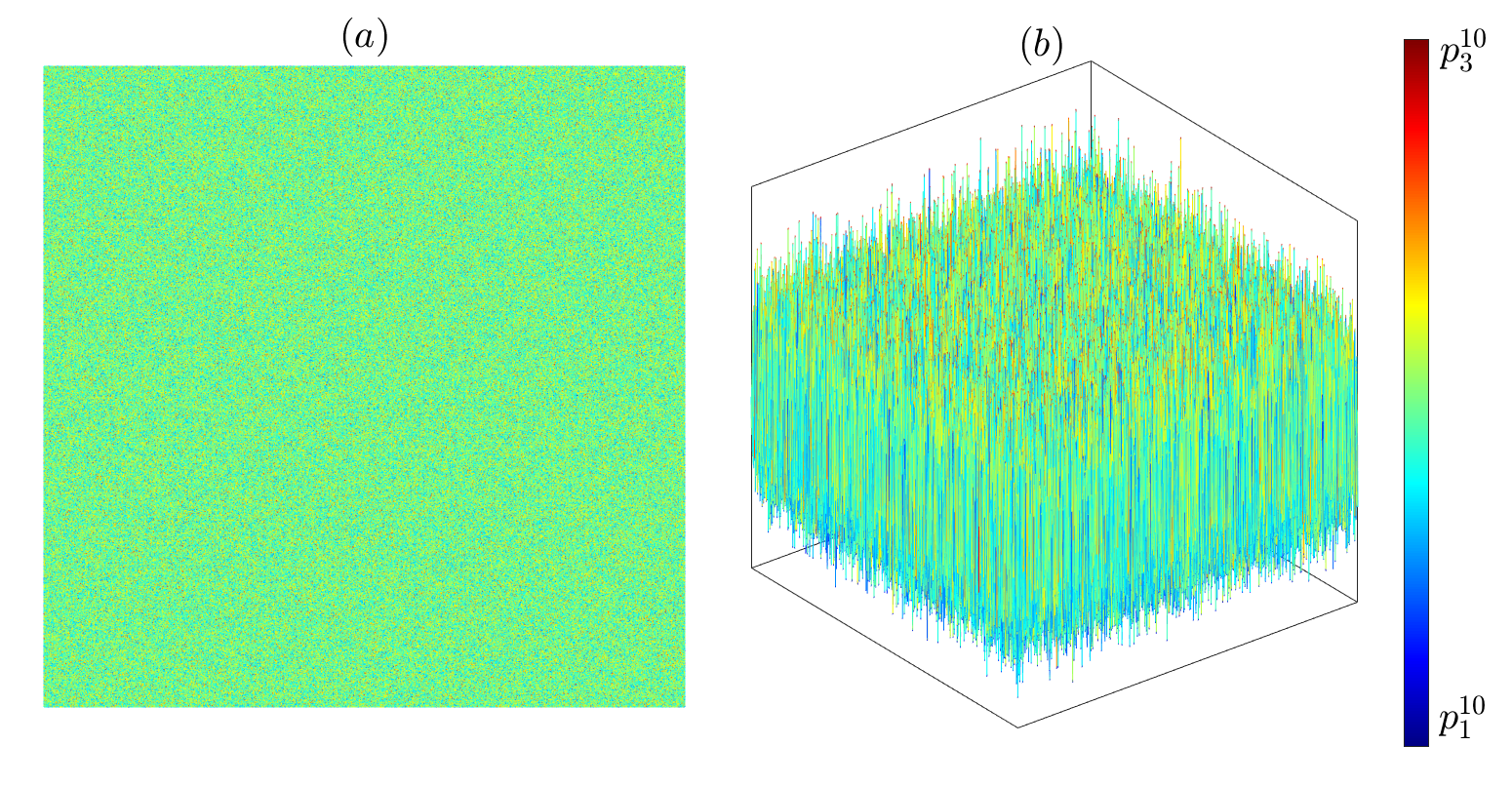}
\caption{(a) Random image obtained by mixing randomly the $\mu$-values of the multifractal surface in Fig. \ref{cuatrinom-2}, (b) 3-dimensional rendering of (a).}
\label{cuatrinomial-random}
\end{figure}

In order to analyze the performance of the proposed method on different types of images, we start by producing two different new ones, built from the same SMM surface of Fig. \ref{cuatrinom-2}. The first one is obtained by reordering the $\mu$-values of the $2^{10}\!\times\!2^{10}$-arrange, from lowest to highest, yielding a complete regular image as shown in Fig. \ref{cuatrinomial-ordered}. The second one, by randomly reordering the same $\mu$-values, yielding a fully noisy image as shown in Fig. \ref{cuatrinomial-random}. These last two images will be also scanned by a Hilbert curve of level 10 and subsequently unfolded into time series.
%%%------------------------------------------------
%%%-------------------------------------
%%%---------------------------------------

For these three time series given by the images, henceforth the multifractal, the ordered and the randomized cases, we compute the values of the corresponding information theory quantifiers according to the methods described in Sec. \ref{sec:methods}, for fixed embedding parameter $D\!=\!6$ and delay parameter $\tau\!=\!1$, to maintain the spatial resolution of the image as much as possible. We also consider the following rigid transformations on each of the original images: $\frac{\pi}{2},\pi$ and $\frac{3\pi}{2}$ angle rotations, and a mirror-symmetrical image, so a set of five images (and corresponding time series) are related to each case.

%%%---------------------------------------------
\begin{figure}[ht!]
\hspace{-1.5cm}
\includegraphics[scale=0.3]{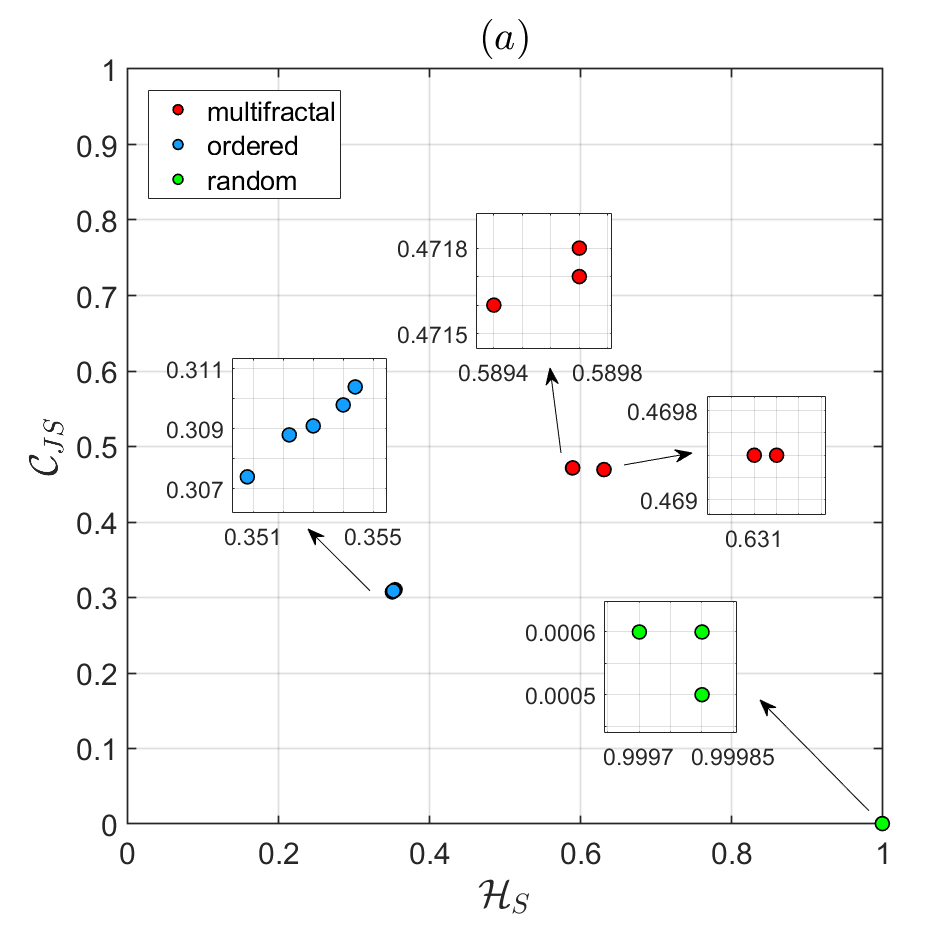}%
\includegraphics[scale=0.3]{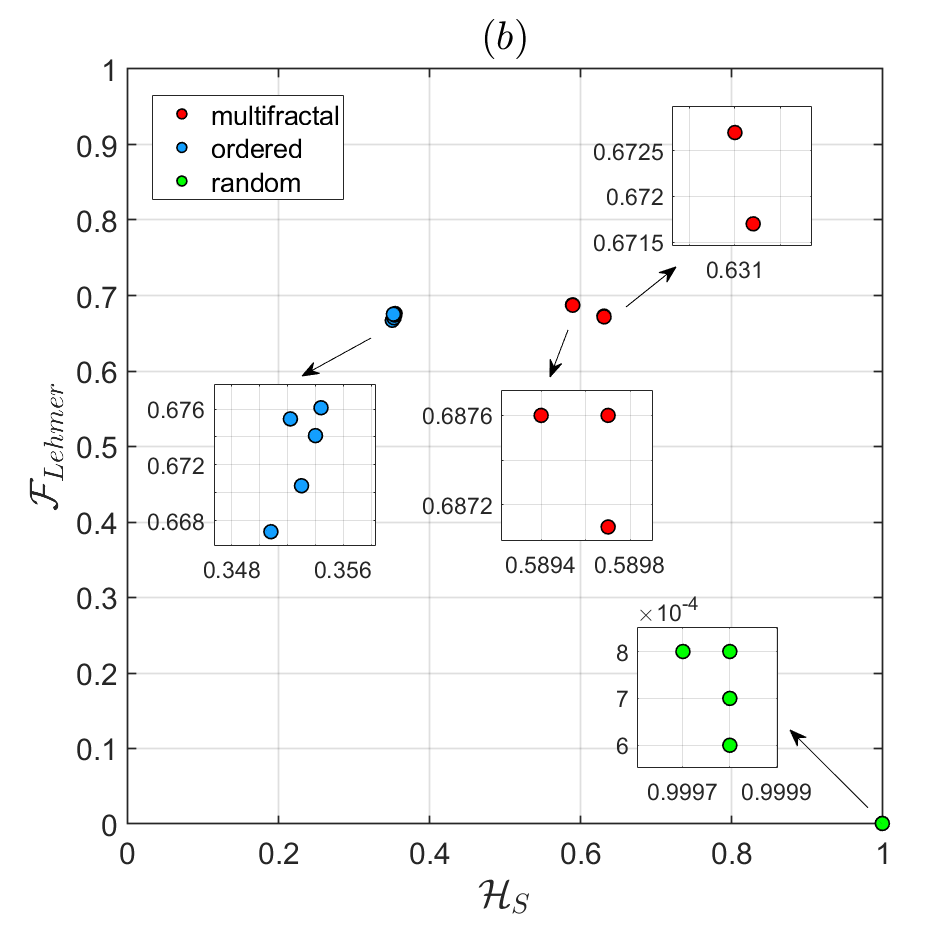}
\caption{CECP (a) and FECP (b) for the SMM (red), the ordered (blue) and the randomized (green) cases. A close up on each group is made to show the true planar locations of the five points corresponding to the original, the rotated ($\frac{\pi}{2},\pi$ and $\frac{3\pi}{2}$) and mirror-symmetrical images.}
\label{cecp-fisher-multif}
\end{figure}

Figures \ref{cecp-fisher-multif}(a) and (b) display the CECP and FECP, respectively, for the fifteen data points. It can be appreciated the net separation in the planar location of the data corresponding to the multifractal, the ordered and the randomized cases, that shows the ability of the purposed method to discriminate the images type between the three classes. The invariance of the results under rotations and mirror-symmetry  transformations is also fully evident, since the points of each group are located almost in the same place. It is necessary to make a close up on each group to be able to distinguish between the five data points, and in the randomized case there are still points in the same place. For the multifractals, there is a very little discrepancy that separates two data points corresponding to the original and the $\pi$ angle rotated images from three data point corresponding to $\frac{\pi}{2}$ and $\frac{3\pi}{2}$ angle rotated and mirror-symmetrical images.

The places of data points are also in accordance to that expected, since the randomized group (green dots) attaches ${\cal H}\!\sim\!1$, ${\cal C}_{JS}\!\sim\!0$ and ${\cal F}\!\sim\!0$, the ordered (blue dots) has the minimun $\cal H$ value, a low ${\cal C}_{JS}$ value and a higher ${\cal F}_{Lehmer}$ value. The multifractal group (red dots) attaches intermediate values of ${\cal H}$ and ${\cal C}_{JS}$ (being the highest of the three groups), and a high ${\cal F}_{Lehmer}$ value similar to the ordered group, which shows the ability of the method to properly capture the rich structuring of these type of images.

%%%==============================================================
\subsection{Brownian surfaces}

In contrast to the deterministic process generating a SMM, the Brownian surfaces are the product of a random process, so the resulting images are closer to real-world ones. They have been introduced by B. Mandelbrot \cite{Mandelbrot} for simulating natural textures, and have since been successfully used to design computer-generated landscapes. 

Here, a random variable, $X(x,y)$, is considered as the height of a surface at each point $(x,y)\!\in\!E$.
For $0\!<\!H\!<\!1$, an {\it index}-$H$ {\it fractional Brownian function} $X\!:\!\mathbb{R}^2\!\rightarrow\!\mathbb{R}$ is a Gaussian random function such that \cite{Falconer}: (i) with probability 1, $X(0,0)\!=\!0$ and $X(x,y)$ is a continuous function of $(x,y)$; (ii) for $(x,y)$, $(h,k)\!\in\!\mathbb{R}^2$, the height increments $X(x+h,y+k)\!-\!X(x,y)$ are normally distributed, with mean 0 and variance $(h^2+k^2)^H\!=\!||(h,k)||^{2H}$. 

\begin{figure}[ht!]
\includegraphics[scale=0.33]{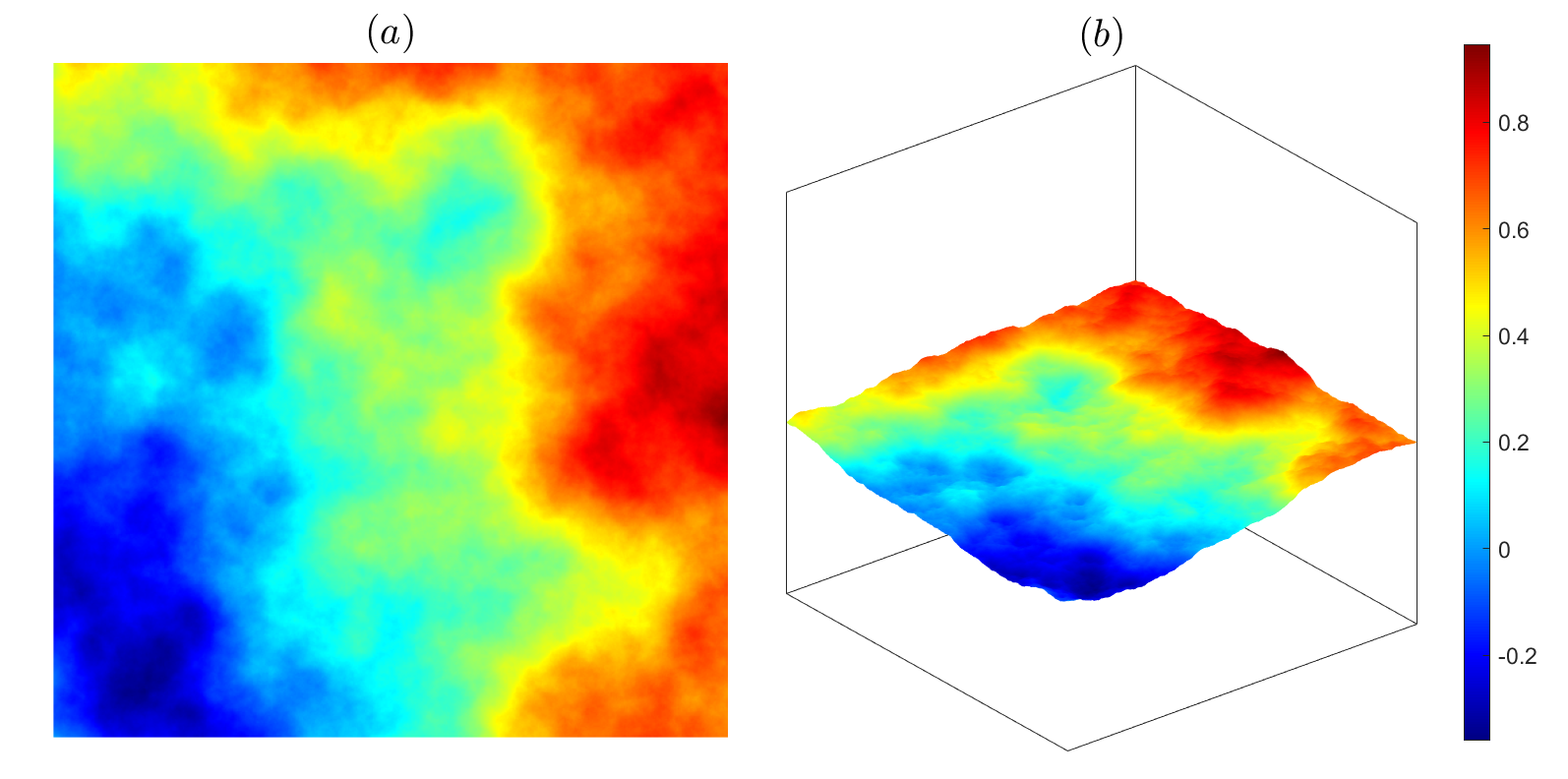}
\caption{(a) Numerical simulation of a $H$-fBs on the unit square for $H\!=\!0.9$, (b) 3-dimensional rendering of (a).}
\label{H-0.9}
\end{figure}

\begin{figure}[ht!]
\includegraphics[scale=0.33]{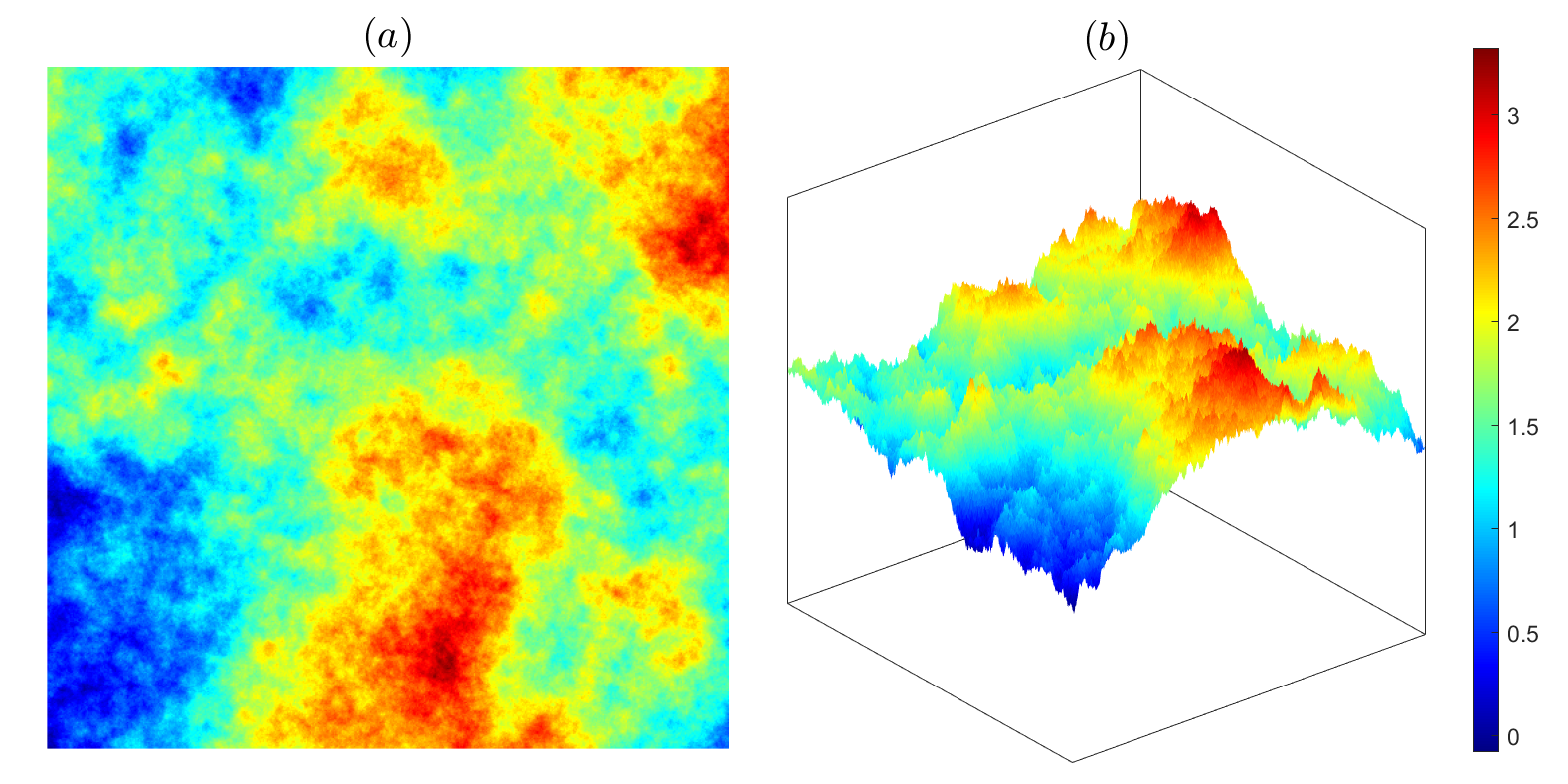}
\caption{(a) Numerical simulation of a $H$-fBs on the unit square for $H\!=\!0.5$, (b) 3-dimensional rendering of (a).}
\label{H-0.5}
\end{figure}

\begin{figure}[ht!]
\includegraphics[scale=0.33]{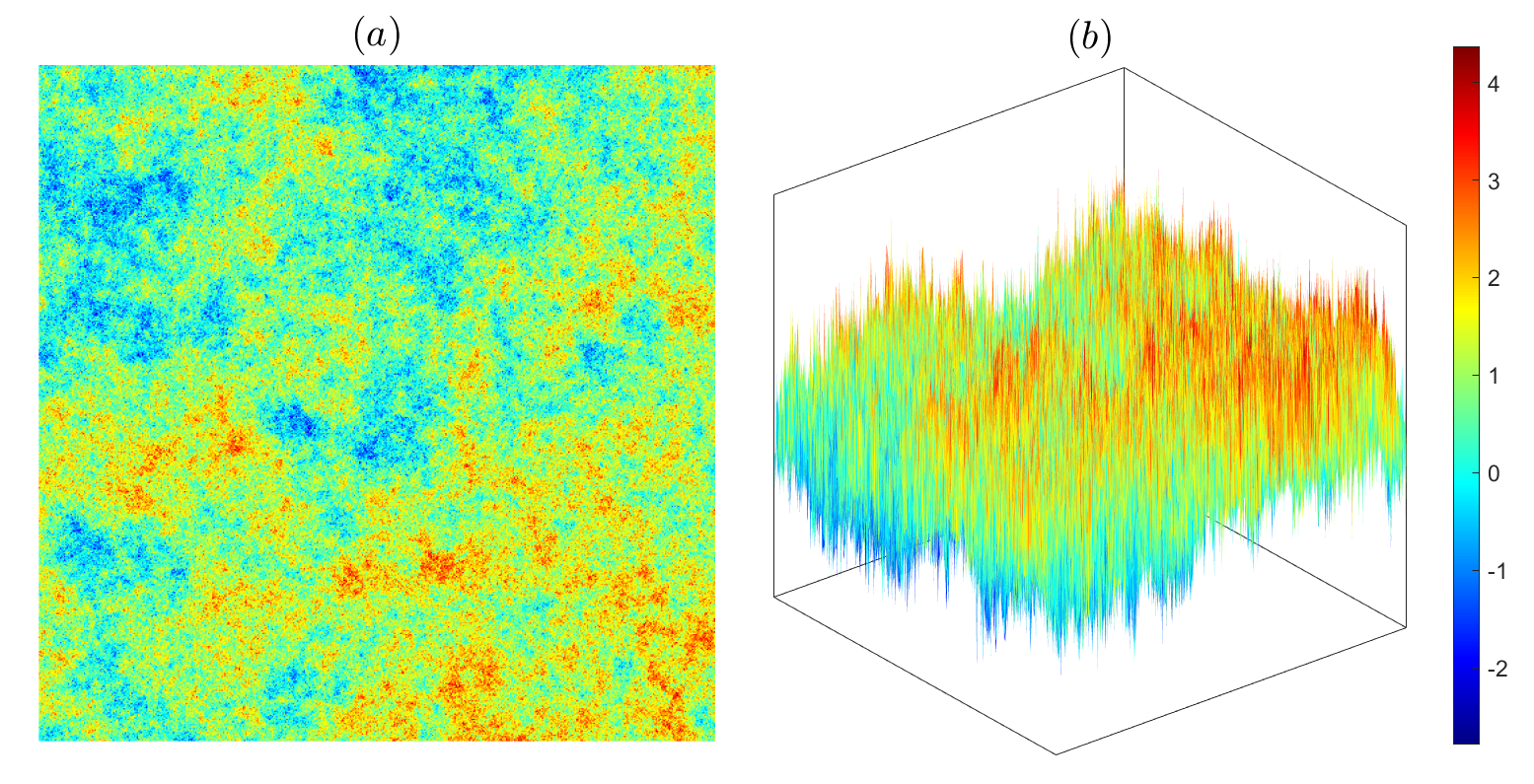}
\caption{(a) Numerical simulation of a $H$-fBs on the unit square for $H\!=\!0.1$, (b) 3-dimensional rendering of (a).}
\label{H-0.1}
\end{figure}

The set $\big\{\big((x,y),X(x,y)\big)\!:(x,y)\!\in\!\mathbb{R}^2\big\}$ is called an {\it index}-$H$ {\it fractional Brownian surface} ($H$-fBs) and it is known that, with probability 1, it has both, Hausdorff and box-counting fractal dimension equal to $3\!-\!H$ \cite{Falconer}. Thus, the closer the $H$ value is to 1, the lower is its fractal dimension and so its level of roughness. On the contrary, the closer the $H$ value is to 0, the higher the fractal dimension and the level of roughness. The index $H$ is also known in the literature as the {\it Hurst exponent} and is indeed used as a roughness parameter. Three numerical simulated examples of $H$-fBs of size $2^{10}\!\times\!2^{10}$ are illustrated in Figures \ref{H-0.9}, \ref{H-0.5} and \ref{H-0.1}, for $H\!=0.9,0.5$ and $0.1$, respectively.
%%%--------------------------------------
%%%-------------------------------------------
%%%---------------------------------------------
%%%----------------------------------------------

%%%--------------------------------------------
\begin{figure}[ht!]
\hspace{-1.3cm}
\includegraphics[scale=0.3]{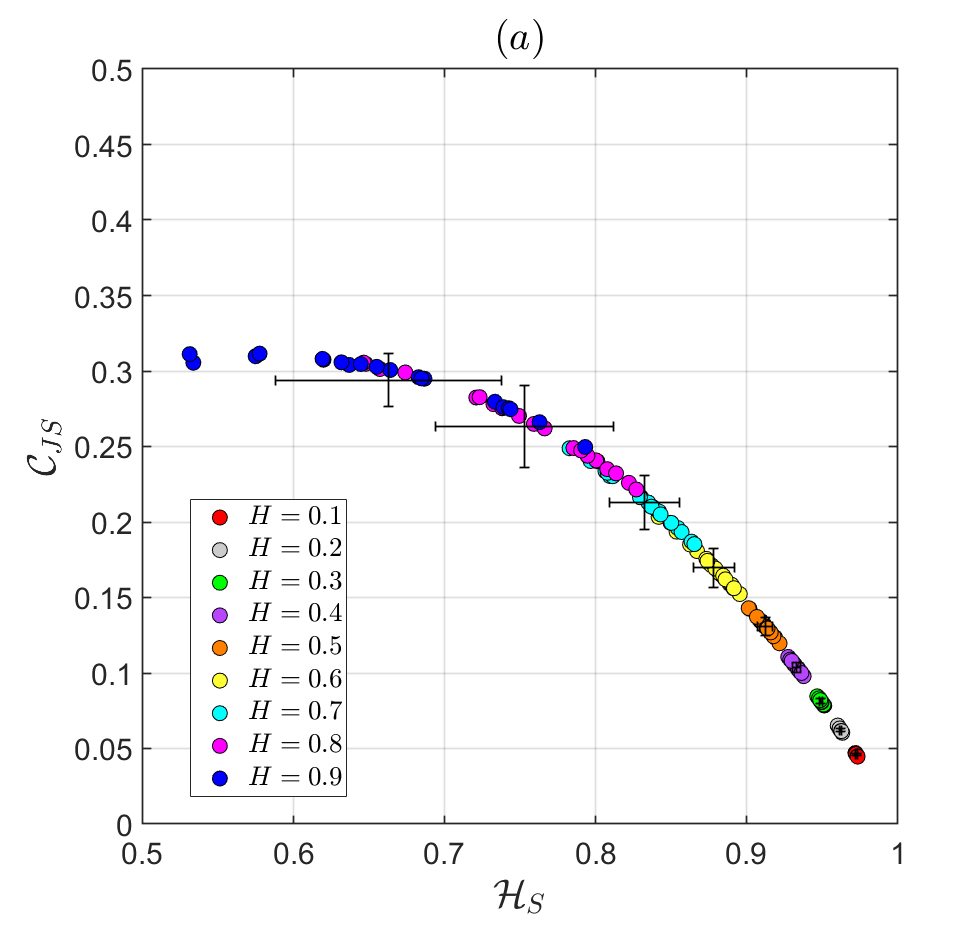}%
\includegraphics[scale=0.3]{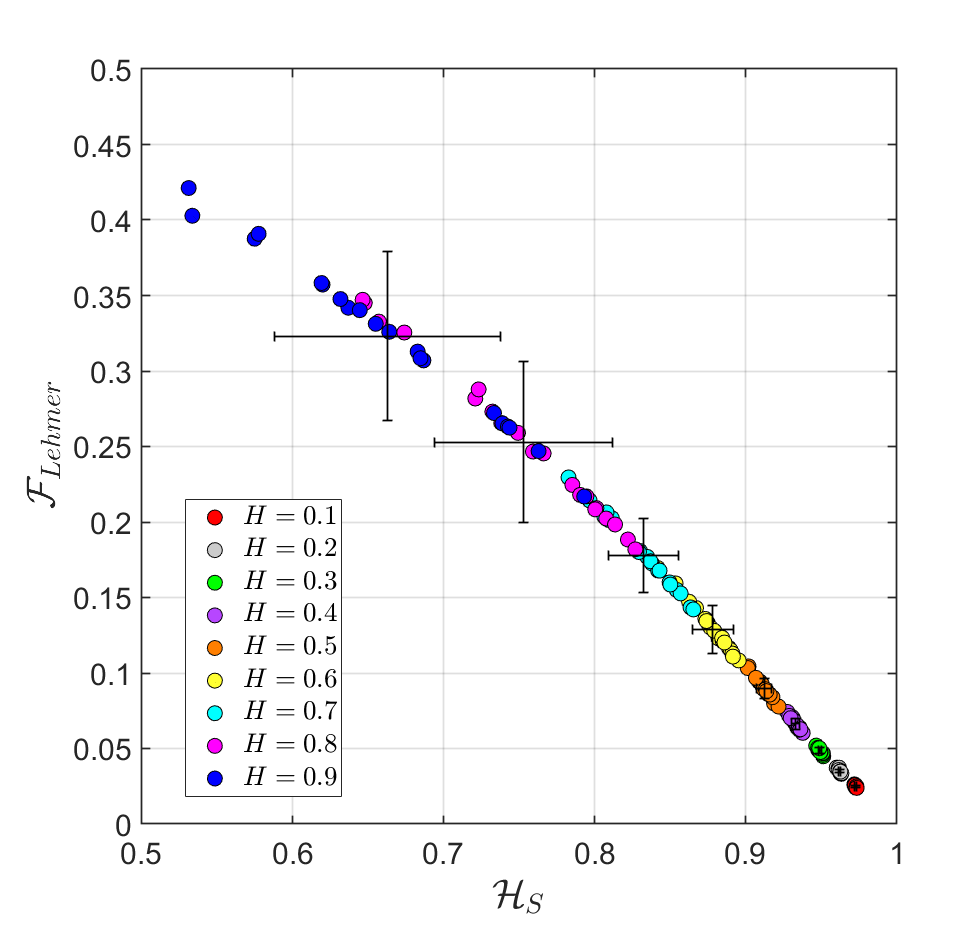}
\caption{CECP (a) and FECP (b) for 30 independent numerical simulations of $H$-fBs for each $H$, $H\!\in\!\{0.1,0.2,\hdots,0.9\}$, for $D\!=\!5$ and $\tau\!=\!1$. Mean and standard deviation for each set of realizations are also plotted as error bars.}
\label{CECP-HFplane-D5-tau1}
\end{figure}
%%%----------------------------------------------

In this work, the information quantifiers involved were calculated for 30 independent simulations for each value of $H$, $H\!\in\!\{0.1,0.2,\hdots,0.9\}$. The results are placed in the CECP and FECP in Fig. \ref{CECP-HFplane-D5-tau1}(a) and (b) respectively, for the parameter values $D\!=\!5$ and $\tau\!=\!1$, together with the mean and standard deviations of each set of realizations. It can be appreciated that both planes discriminate well the groups taking into account the random component that all images have. The data points in the CECP fall in a curve according to the expected values of ${\cal H}\!\sim\!1$ and ${\cal C}\!\sim\!0$  for negative correlated surfaces ($H\!<\!0.5$) whose images appear more  ``desordered'', while for positive correlated ($H\!>\!0.5$) the entropy ${\cal H}$ decreases and the complexity ${\cal C}_{JS}$  increases as $H$ scales to $0.9$. In the FECP instead, the data points locate in a straight line revealing what appears to be an inverse linear relationship between $\cal H$ and $\cal F$. Note that $\cal F$ can reach values higher than $\cal C$ ones for $H\!=\!0.8,0.9$ and, on the contrary, lower values for $H\!=\!0.1, 0.2$. This is because, as a local quantifier, $\cal F$ is more sensitive to capture local correlations just characterized by the $H$ exponent. This illustrates the relevance to add the FECP to the set of tools for studying image complexity.

%%%----------------------------------------------
\begin{figure}[ht!]
\hspace{-1.3cm}
\includegraphics[scale=0.3]{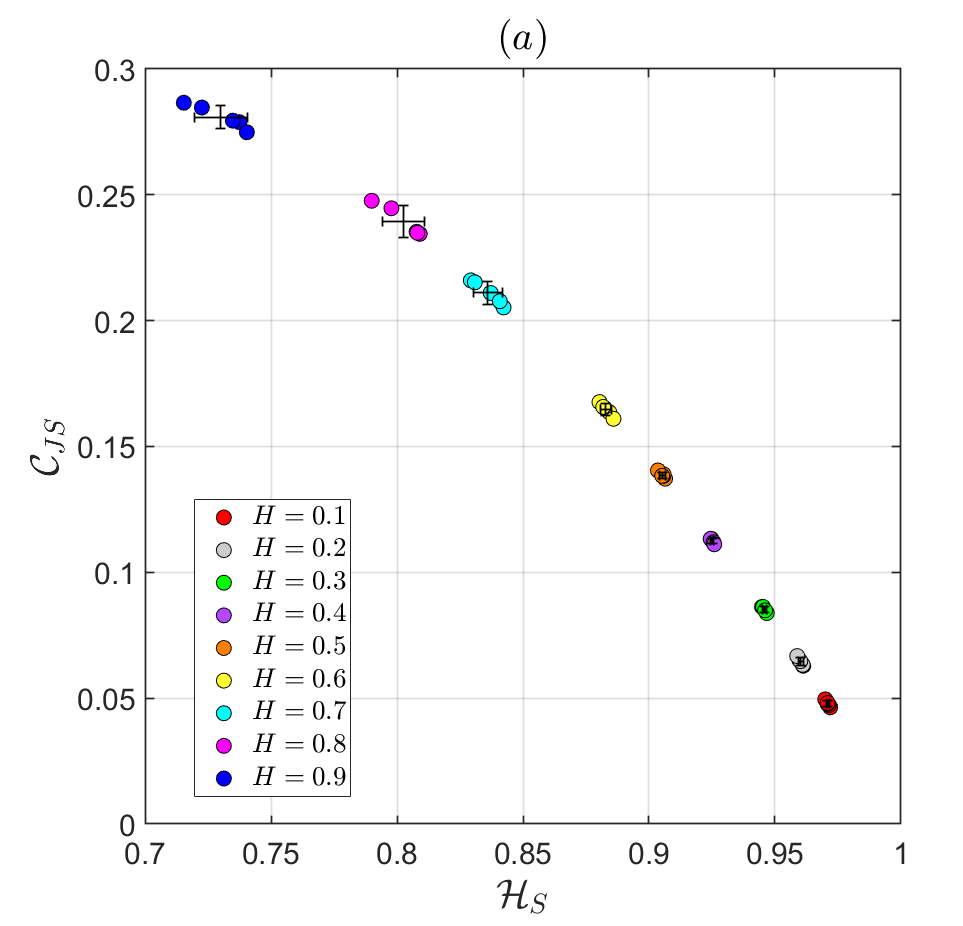}%
\includegraphics[scale=0.3]{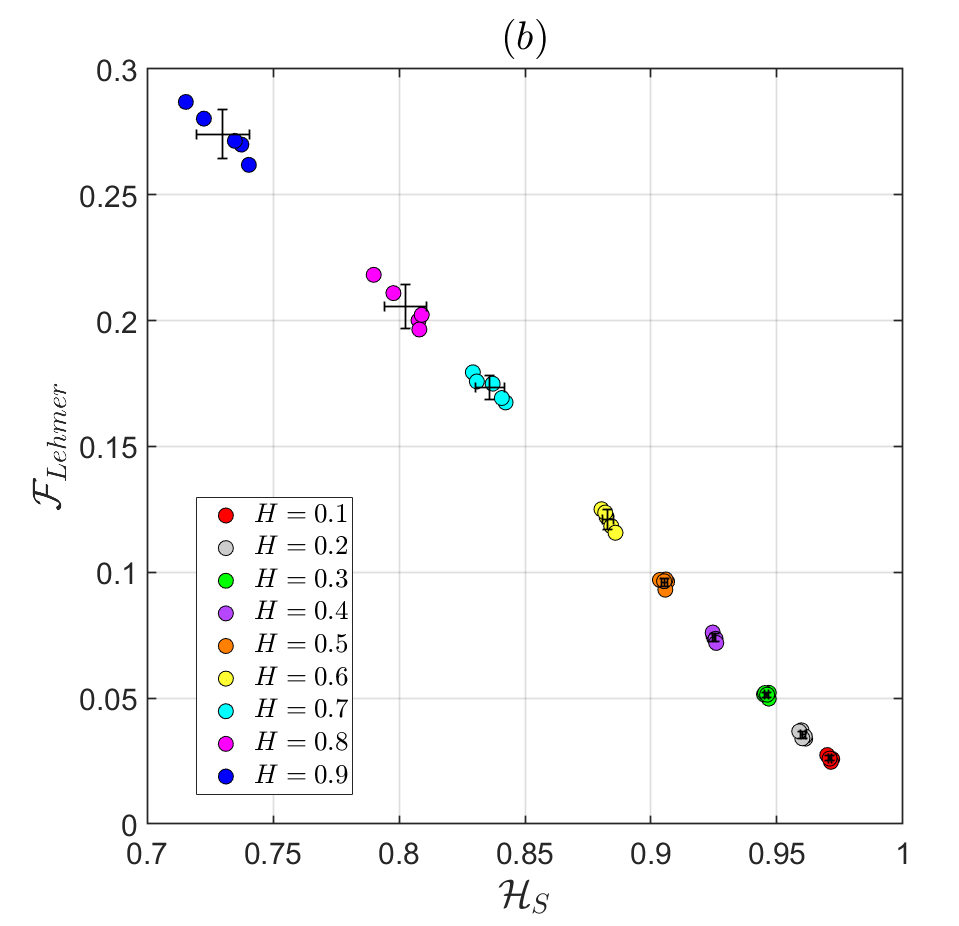}
\caption{CECP (a) and FECP (b) for a single simulation for each $H$ value in Fig. \ref{CECP-HFplane-D5-tau1} along with the corresponding rotated ($\frac{\pi}{2},\pi$ and $\frac{3\pi}{2}$) and mirror-symmetrical images, $D\!=\!5$, $\tau\!=\!1$.}
\label{CECP-HFplane-rotac-D5-tau1}
\end{figure}
%%%--------------------------------

Figures \ref{CECP-HFplane-rotac-D5-tau1}(a) and (b) depict the CECP and FECP respectively, for a single realization of each $H$ value in Fig. \ref{CECP-HFplane-D5-tau1} along with their rotations at $\frac{\pi}{2}, \pi$ and $\frac{3\pi}{2}$, and the mirror-symmetrical image. The invariance of the results against those transformations is here also outright.

%%%======================================================
\subsection{Normalized Brodatz}

\begin{figure}[ht!]
\subfloat[D15]{
\includegraphics[width=0.25\textwidth]{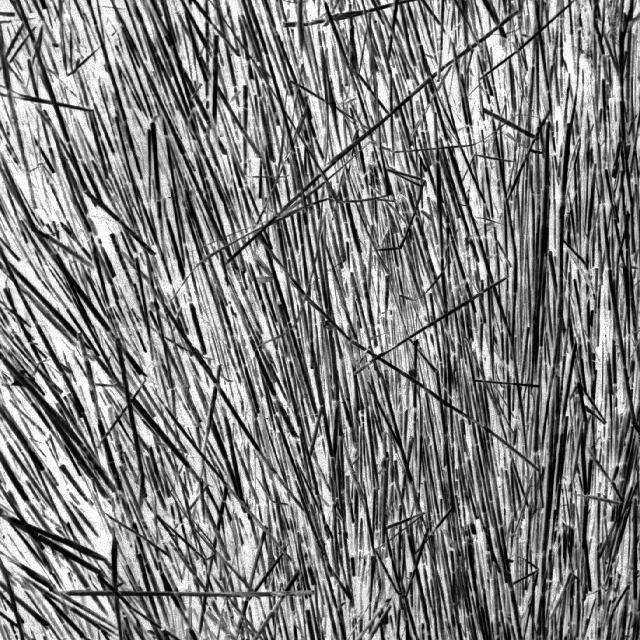}}
  \subfloat[D16]{
\includegraphics[width=0.25\textwidth]{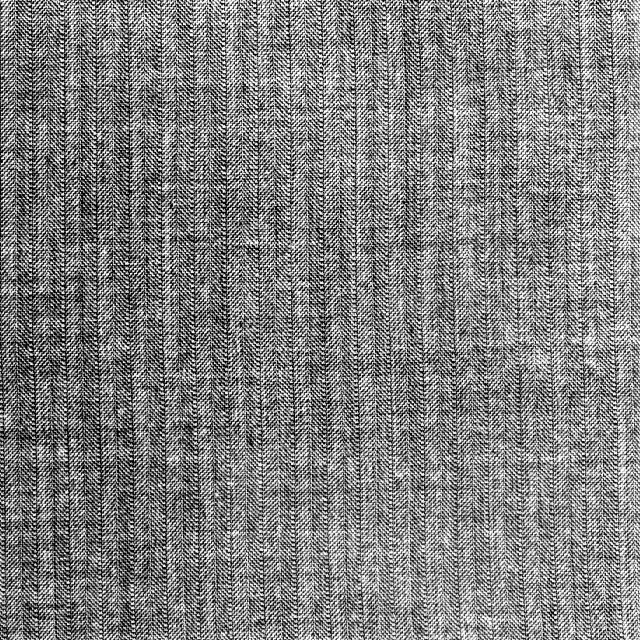}
 }
  \subfloat[D44]{
   \label{f:D44}    \includegraphics[width=0.25\textwidth]{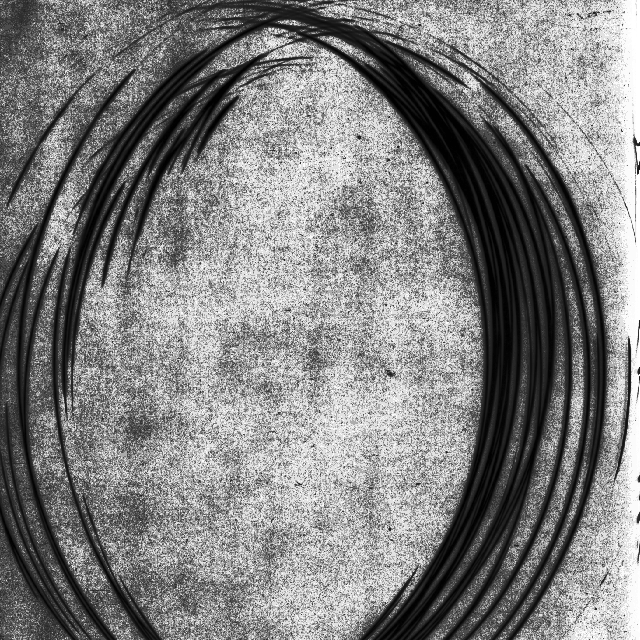}}
    \subfloat[D49]{
   \label{f:D49}    \includegraphics[width=0.25\textwidth]{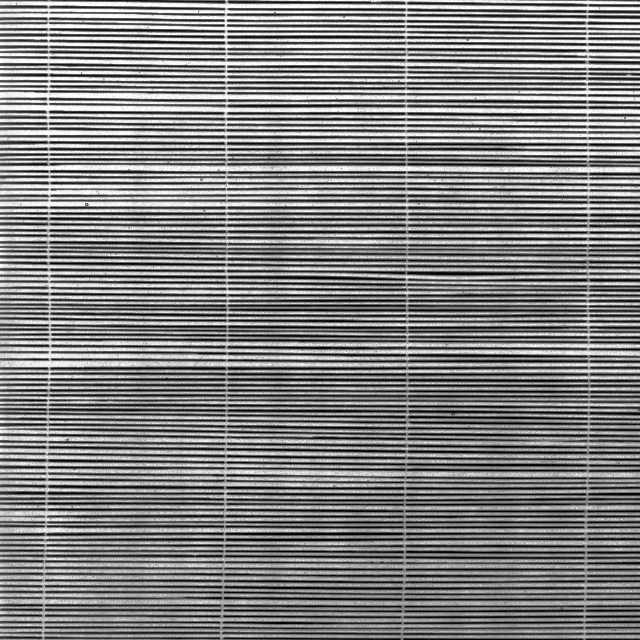}} \\
\subfloat[D71]{
   \label{f:D71}    \includegraphics[width=0.25\textwidth]{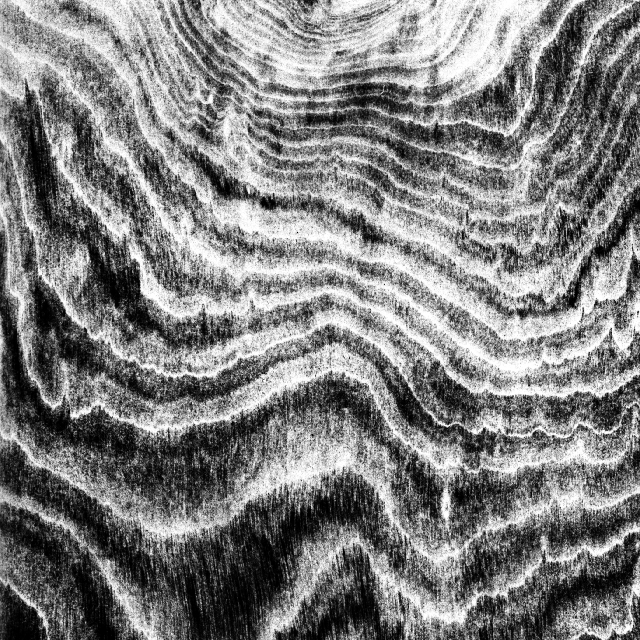}}
\subfloat[D93]{
   \label{f:D93}    \includegraphics[width=0.25\textwidth]{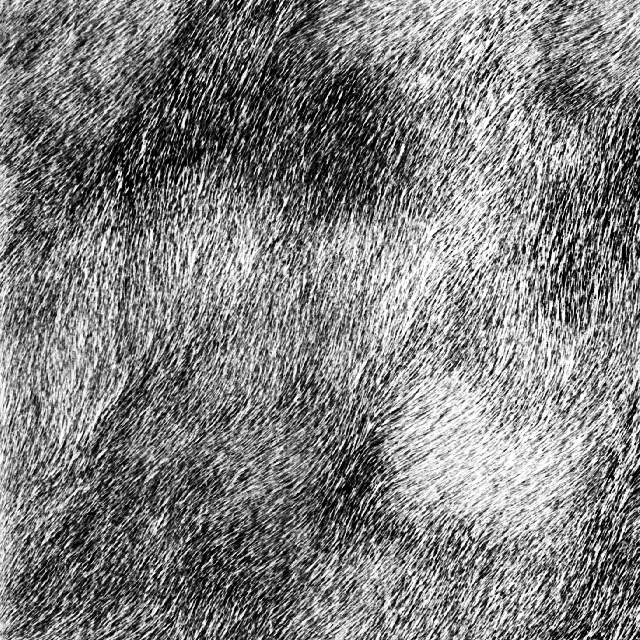}}  
\subfloat[D101]{
   \label{f:D101}    \includegraphics[width=0.25\textwidth]{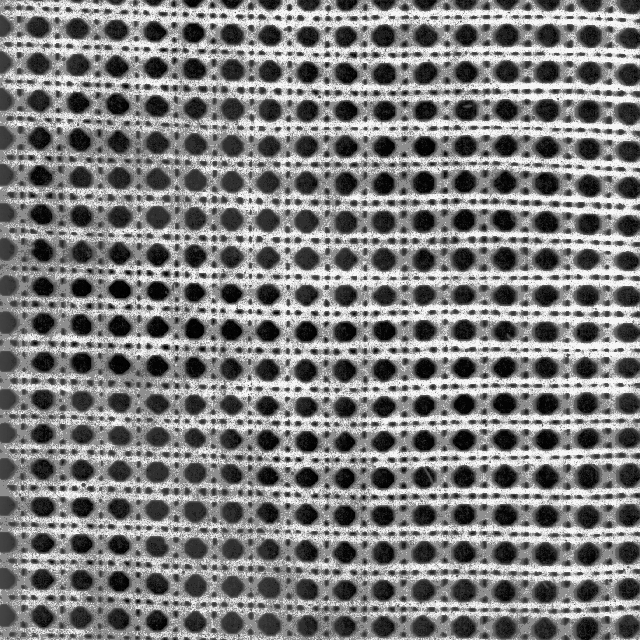}}
\subfloat[D102]{
   \label{f:D102}    \includegraphics[width=0.25\textwidth]{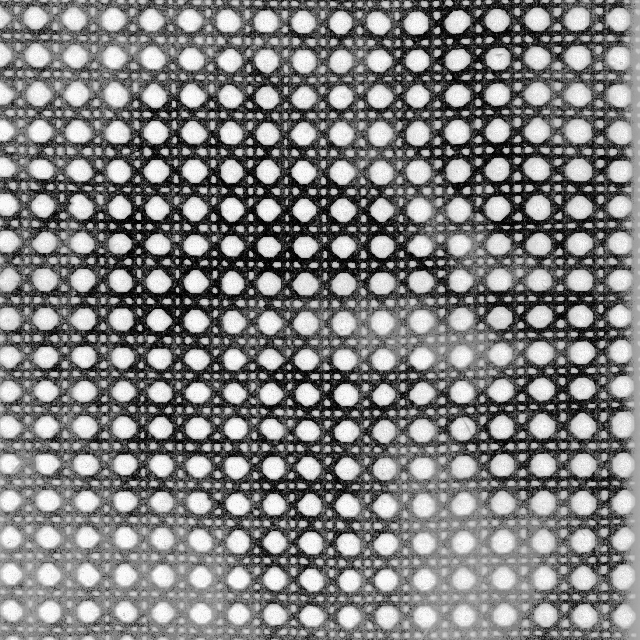}}
 \caption{Eight Normalized Brodatz  \tc{Blue}{Textures selected from NBT database with their corresponding labels.}}
 \label{f:Normal-Brodatz}
\end{figure}

We apply the method to a real-world surface dataset by taking images from the Normalized Brodatz Texture database (NBT) \citep{AbdelmounaimeDong2013}\footnote{ The images are freely available from: \url{https://multibandtexture.recherche.usherbrooke.ca/normalized_brodatz.html}}. This database comprises 112 photos from the album ‘Textures: A Photographic Album for Artists and Designers’ (\cite{Brodatz1966}).  
The original Brodatz images have different background intensities and some of these texture images could be discriminated using only this feature (i.e., first-order statistics) without using texture information. Thus, they were enhanced by a normalization process which removed the greyscale background effect. Figure \ref{f:Normal-Brodatz} shows 8 different images from the NBT database, selected as follow-up examples of the various techniques we implemented throughout the work. In addition, some of them coincide in having been analyzed in other works, such as \cite{ZUNINO2010efficiency}, for a possible confrontation. The images of the NBT database are $640\!\times\!640$ pixels in size and use 8 bits per pixel, allowing for 256 different shades of gray. They were cropped to a size of $512\!\times\!512$ in order to be scanned by a Hilbert curve of level 9. 

%\tc{red}{salvo 102 y 16, las demás las menciona Zunino, se corresponden a distintas posiciones en la curva entropía complejidad.}

%%%--------------------
\begin{figure}[ht!]
\hspace{-1.5cm}
\includegraphics[scale=0.3]{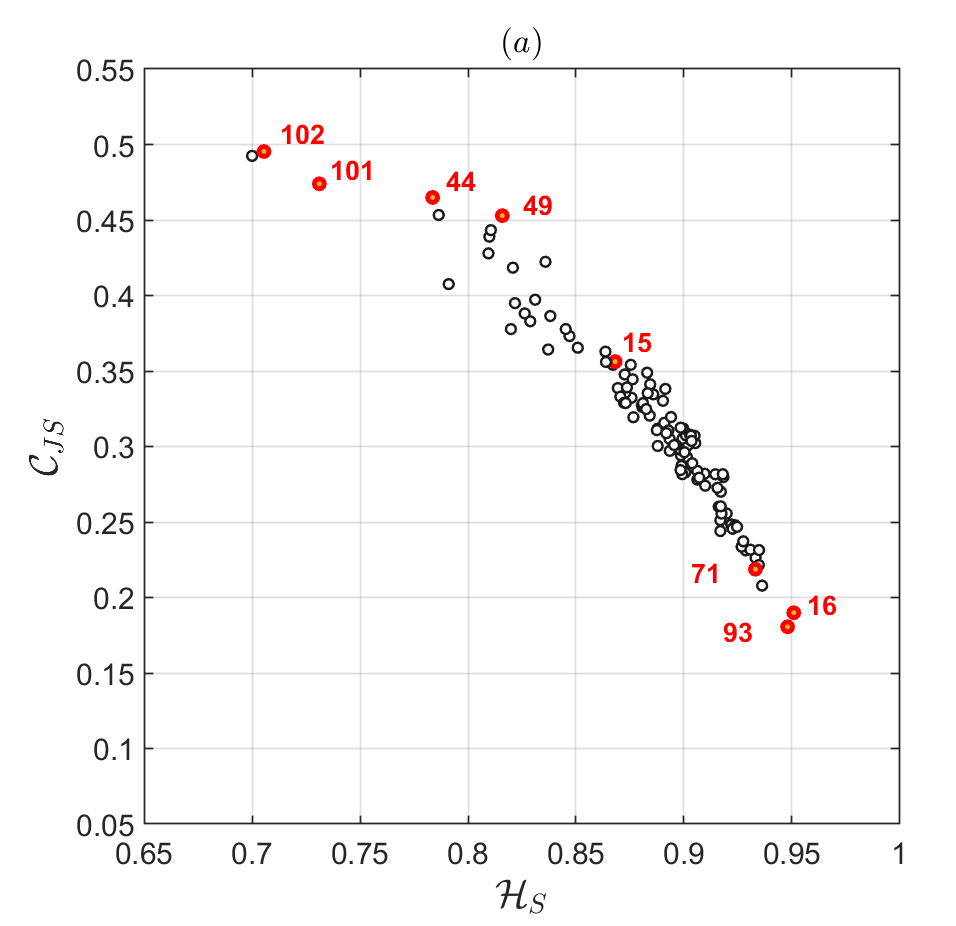}%
\includegraphics[scale=0.3]{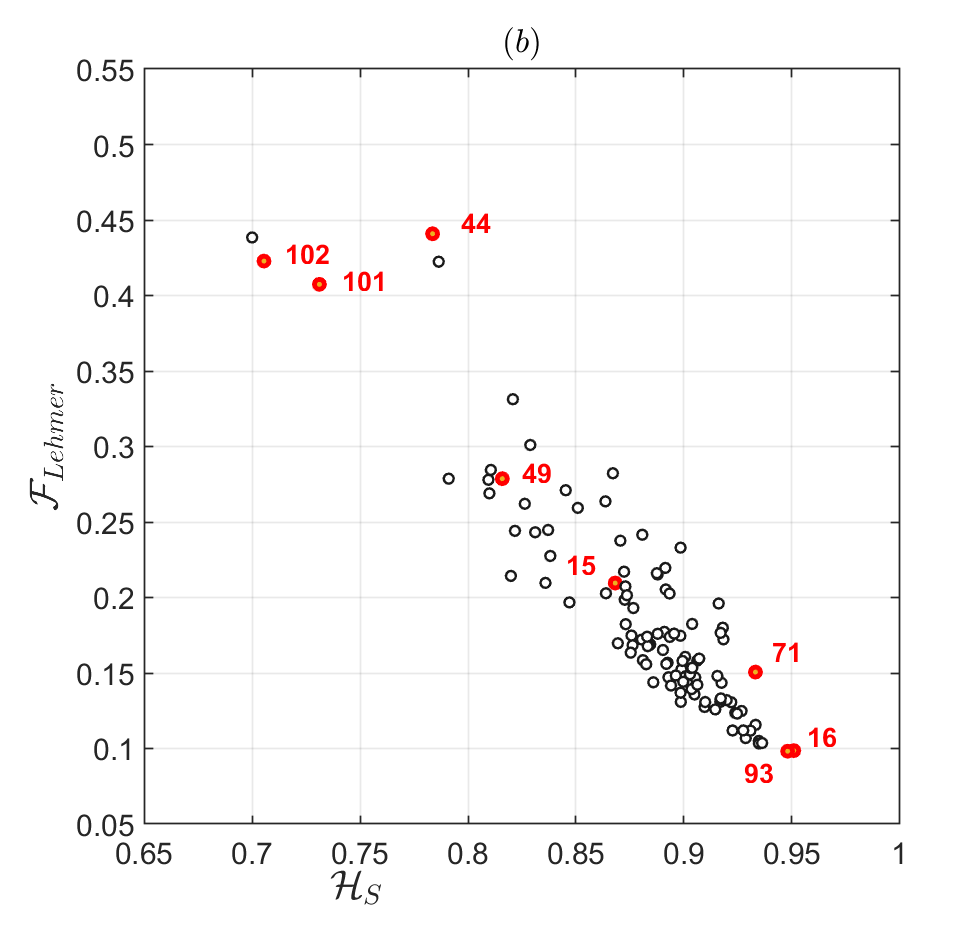}
\caption{CECP (a) and FECP (b) for the Normalized Brodatz set with $D\!=\!8$ and $\tau\! =\! 1$. The data points and labels of the eight selected textures (Fig. \ref{f:Normal-Brodatz}) are highlighted in red.}
\label{f:Normal-Brodatz-Entropy-Complexity}
\end{figure}

%%%----------------------------------------------

Both planar representations, CECP and FECP, were calculated for the 112 images of the database. Figure \ref{f:Normal-Brodatz-Entropy-Complexity}  illustrates the locations in both planes for the case $D\!=\!8$ and $\tau\!=\!1$, where the data points corresponding to  the 8 selected images are highlighted in red.
It can be seen that they spread over the suggested representation planes, but the FECP discriminates the images better. Thus, it's crucial to calculate both quantifiers. Regarding the selected textures in the FECP, for example, one can distinguish in Fig. \ref{f:Normal-Brodatz-Entropy-Complexity},  three groups according to their FIM values: D16, D71 and D93 (the lowest),  D15 and D49 (the medium ones) and D44, D101 and D102 (the highest). Looking at these groups in Fig. \ref{f:Normal-Brodatz}, it is possible to relate them with different type of patterns: D44, D101 and D102 have a regular design with certain  symmetry around a central point. Besides D101 and D102 display a periodic motif and so they have a higher value of entropy $\cal H$ than D44. D15 and D49 seems to have a preferential direction of plot drawing, vertical and horizontal, respectively, being sharper in D49, and that is why its entropy is greater than that of D15. Finally, D71 and D93 are more ``disordered'' designs and D16 has almost no contrasts being therefore close to the uniform distribution. 
%%%--------------------------------
\begin{figure}[h!]
\centering
\includegraphics[scale=0.35]{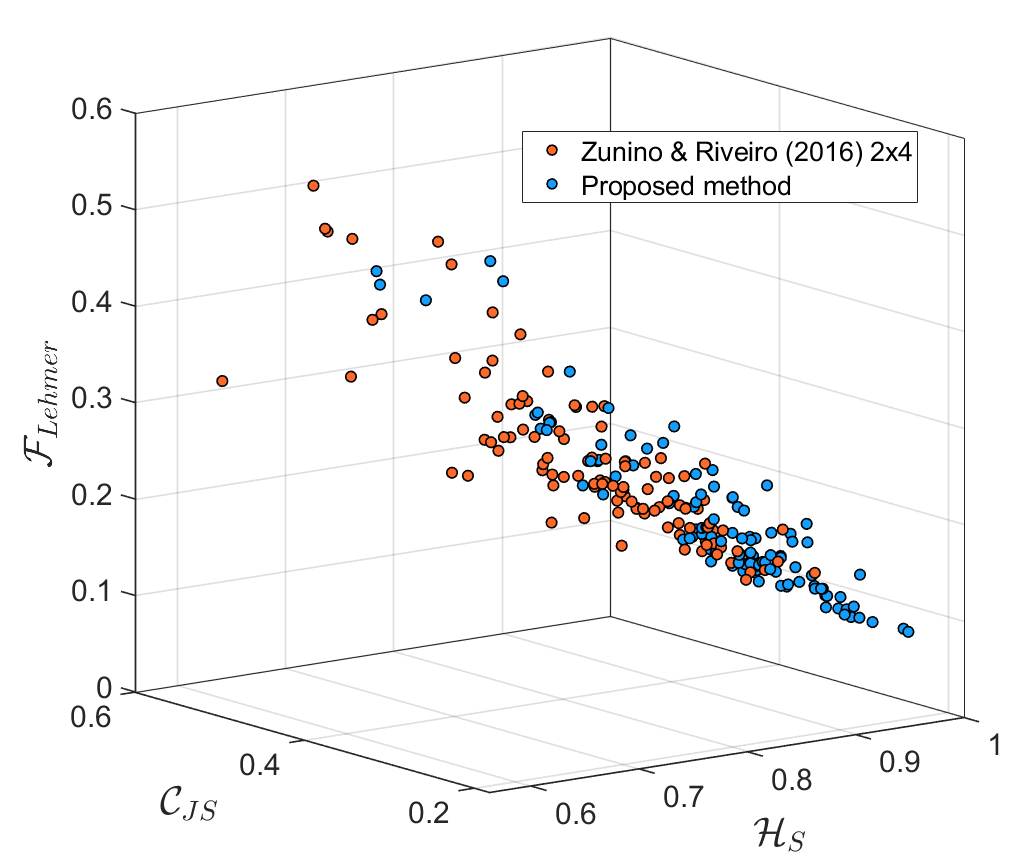}%
\caption{Entropy-Complexity-Fisher space of Normalized Brodatz Texture database. Embedding dimensions $D_x\!\times\!D_y\!=\!2\!\times\!4$ and delays $\tau_x\!=\!1$, $\tau_y\!=\!1$ for Zunino \& Ribeiro method (\cite{ZUNINO2016679}) (red points); $D\!=\!8$ and $\tau\!=\!1$ for the proposed method (blue points).}
\label{NosVsZunino}
\end{figure}

In order to compare our results with those by  Zunino \& Rivero \cite{ZUNINO2016679}, we propose a three dimensional representation of the information quantifiers. For each $P$ the triple $\big({\cal H}[P],{\cal C}_{JS}[P],{\cal F}[P]\big)$ is located in the ${\cal H}\!\times\!{\cal C}\!\times\!{\cal F}$ space. We also compute the FIM for the 112 pictures of the NBD, following the row-wise order used in \cite{ZUNINO2016679} for scanning the images when computing the statistical complexity, by using here submatrices of embedding dimensions $D_x\!\times\!D_y\!=\!2\!\times\!4$ and delays $\tau_x\!=\!1$, $\tau_y\!=\!1$. For that aim we used an embedding dimension 
$D\!=\!8$ and $\tau\!=\!1$ when applying the Hilbert curve method (our proposal). The results are depicted in
Fig. \ref{NosVsZunino}. We observe that our approach reaches higher values for the permutation entropy and begins with lower values of statistical complexity and FIM. This figure shows that adding the FIM is useful to enhance image discrimination\footnote{ We also computed the quantifiers for $D_x\!\times\!D_y\!=\!4\!\times\!2$ for comparison, but since the results were very similar, we decided to represent only the case $2\!\times\!4$ for simplicity.}

%%%%%%%%===========================================
\subsection{Colored Brodatz}

\begin{figure}[ht!]
 \centering
  \subfloat[D15]{
   \label{f:D15-C}
    \includegraphics[width=0.25\textwidth]{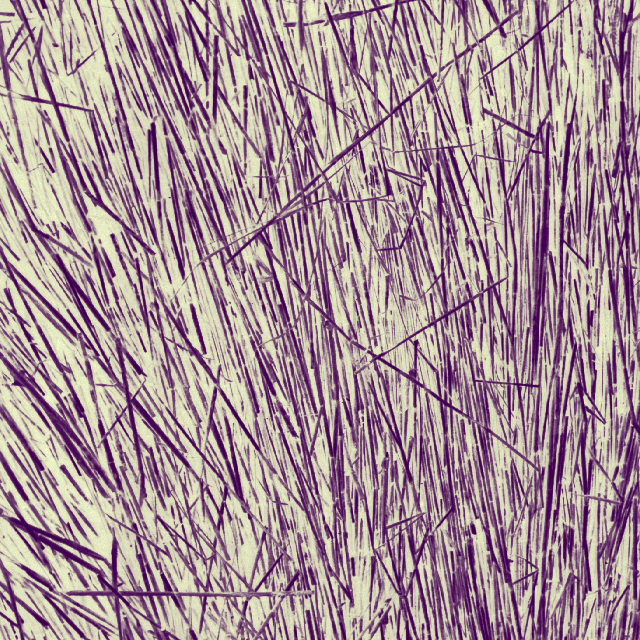}}
  \subfloat[D16]{
   \label{f:D16-C}
    \includegraphics[width=0.25\textwidth]{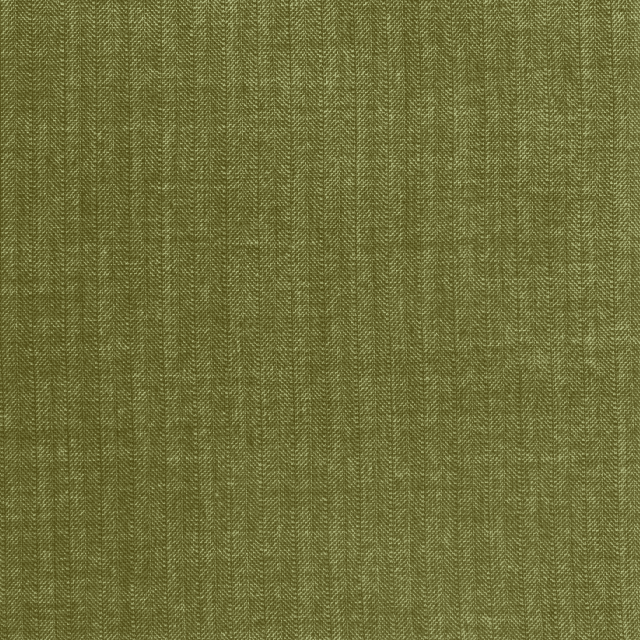}}
  \subfloat[D44]{
   \label{f:D44-C}
    \includegraphics[width=0.25\textwidth]{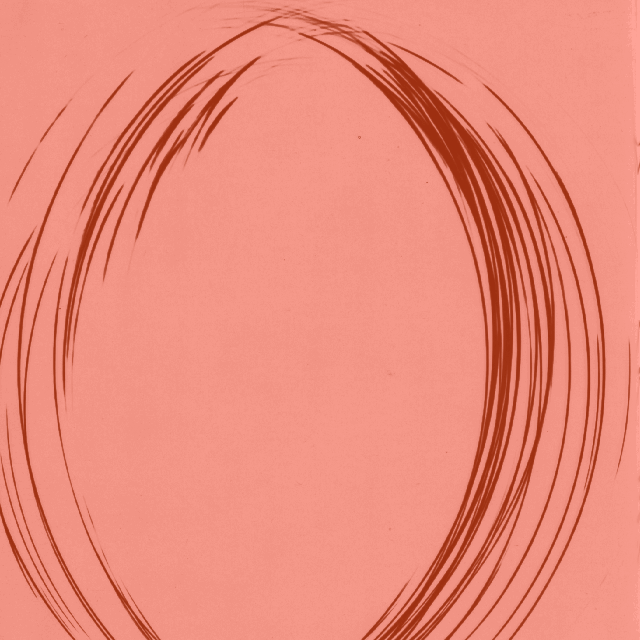}}
    \subfloat[D49]{
   \label{f:D49-C}
    \includegraphics[width=0.25\textwidth]{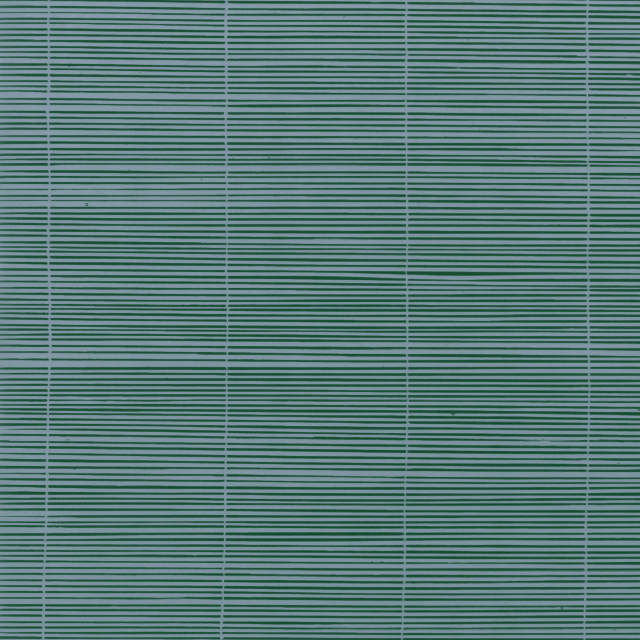}} \\
\subfloat[D71]{
   \label{f:D71-C}
    \includegraphics[width=0.25\textwidth]{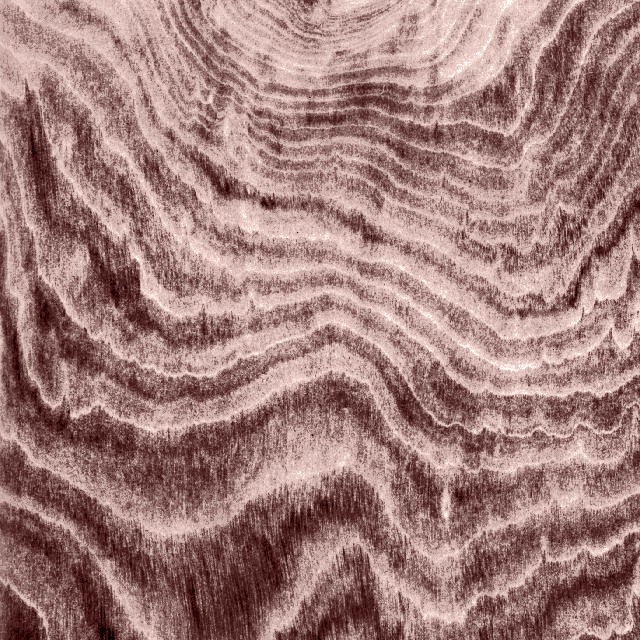}}
\subfloat[D93]{
   \label{f:D93-C}
    \includegraphics[width=0.25\textwidth]{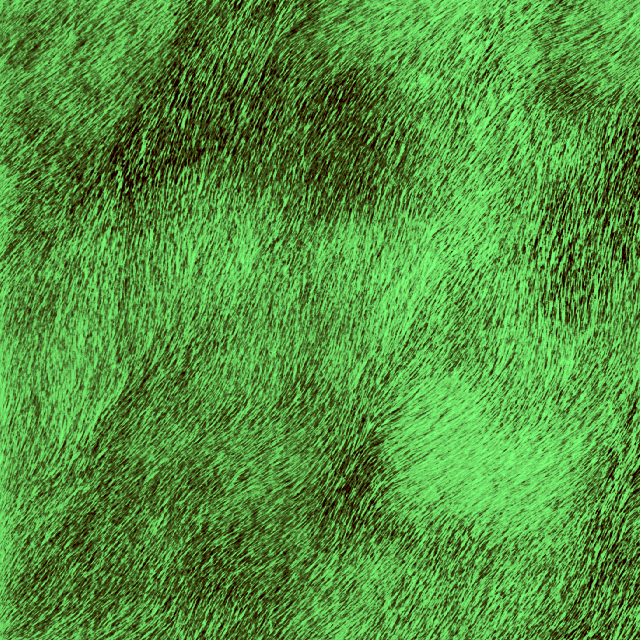}}    
\subfloat[D101]{
   \label{f:D101-C}
    \includegraphics[width=0.25\textwidth]{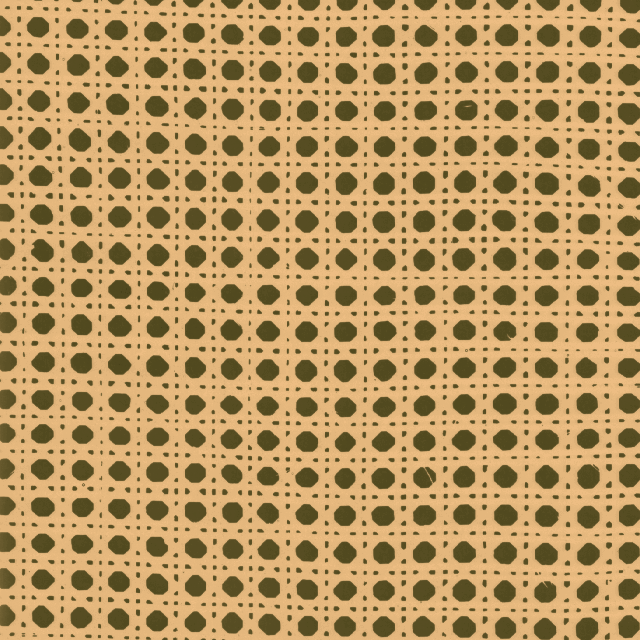}}
\subfloat[D102]{
   \label{f:D102-C}
    \includegraphics[width=0.25\textwidth]{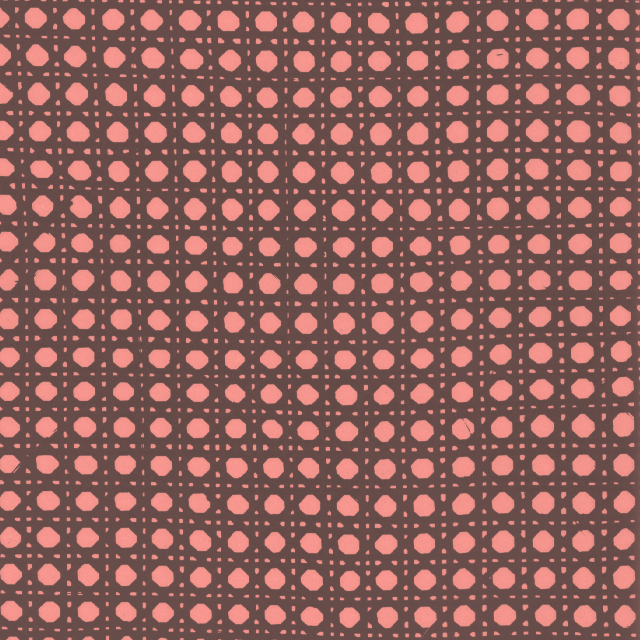}}
 \caption{Colored Brodatz Textures selected from CBT database with their corresponding labels. They are colored versions of the textures in Fig. \ref{f:Normal-Brodatz}.}
 \label{f:Colored-Brodatz}
\end{figure}

\begin{figure}
\hspace{-1.5cm}
    \includegraphics[scale=0.3]{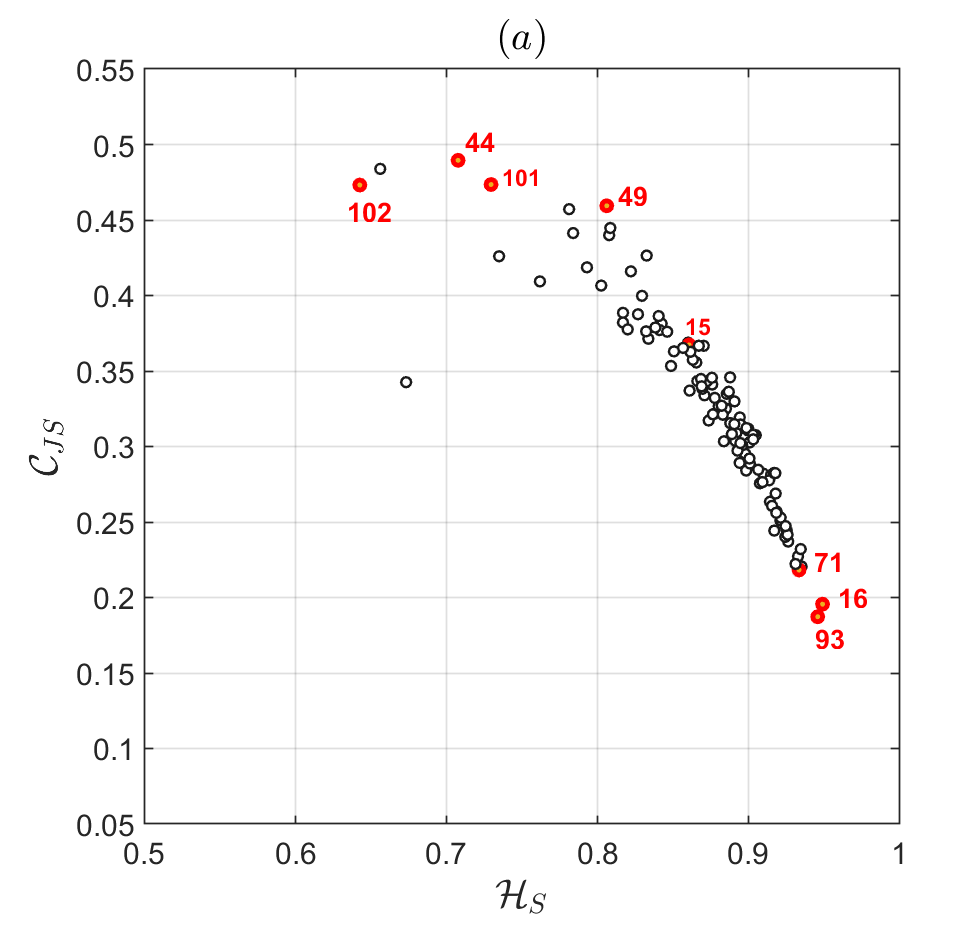}%
    \includegraphics[scale=0.3]{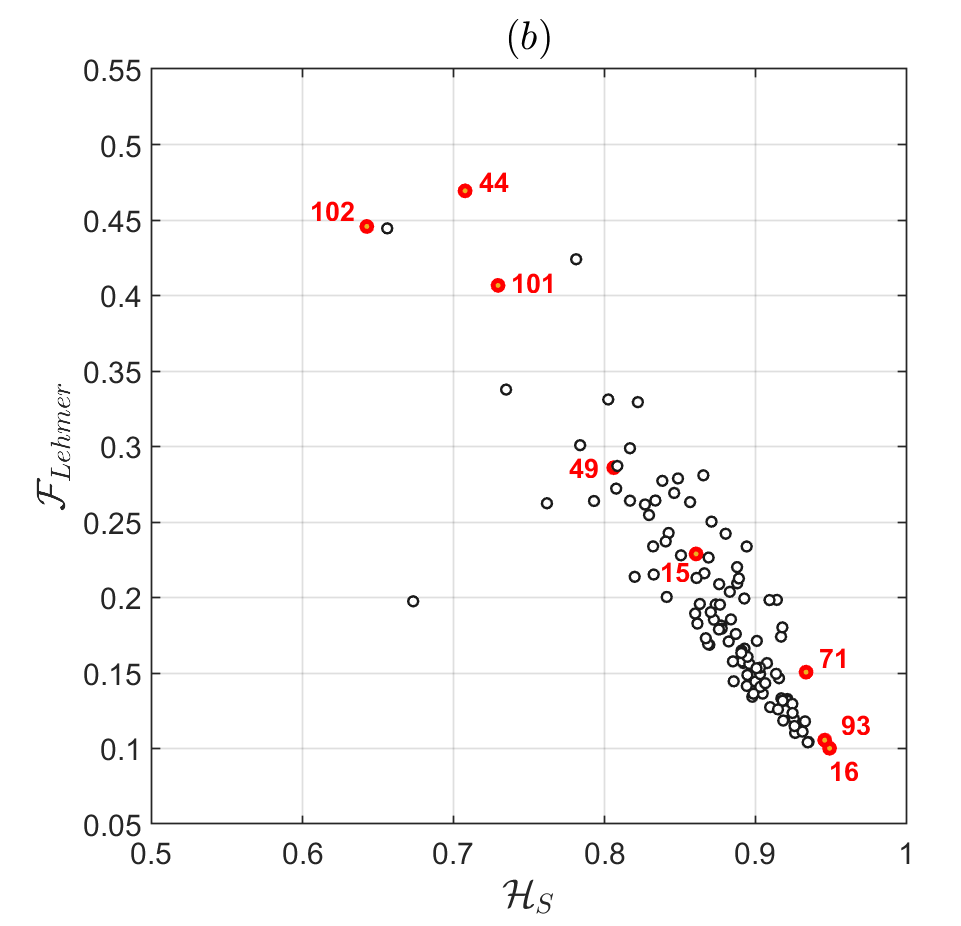}
    \caption{Colored Brodatz set with with $D\!=\!8$, $\tau\!=\!1$ (a) CECP (b) EFCP.}
    \label{f:Colored-Brodatz-Entropy-Complexity}
\end{figure}

In \cite{AbdelmounaimeDong2013}, it is also offered a Colored Brodatz Texture database (CBT), which consists of a colored version of the 112 Brodatz images, and has the advantage of retaining the rich textural content of the originals while having a wide variety of colors. Figure \ref{f:Colored-Brodatz} shows the colored versions of the 8 selected pictures in Figure \ref{f:Normal-Brodatz}. We perform the proposed method to these new images and the results are depicted in Fig. \ref{f:Colored-Brodatz-Entropy-Complexity}. Comparing the latter with Fig. \ref{f:Normal-Brodatz} it can be seen that the location of the normalized and colored textures in the CECP and FECP representations do not differ significantly, as highlighted with the 8 pictures selected for this comparison. The only exception in the color version is one of the images (labeled D111 in both databases), which behaves as an outlier, probably due to over-coloring effect. Therefore, the proposed classification method could be applied either to black and white or to color images. 

%%%%%%%%%%%%%%%%%%%%%%%%%%%%%%%%%%%%%%%%%%%%%%%%%%%%%%
\subsection{Rotated Brodatz}

\begin{figure}[ht!]
 \centering
  \subfloat[0°]{
       \includegraphics[width=0.25\textwidth]{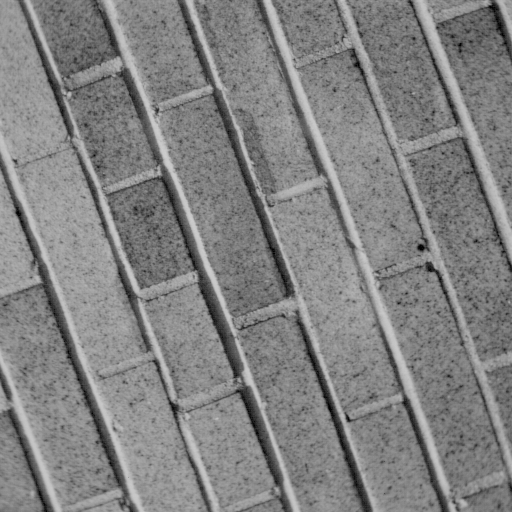}}
  \subfloat[30°]{
      \includegraphics[width=0.25\textwidth]{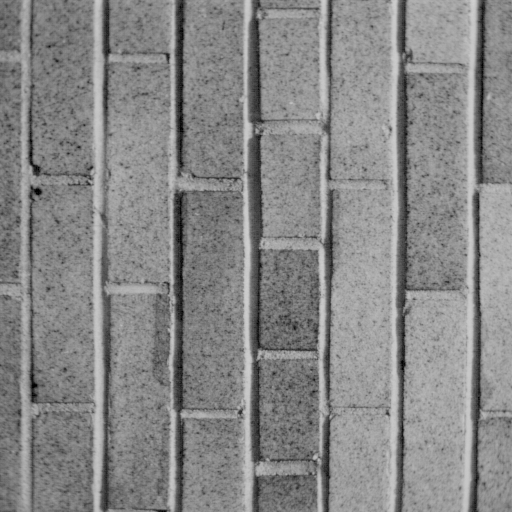}}
  \subfloat[60°]{
      \includegraphics[width=0.25\textwidth]{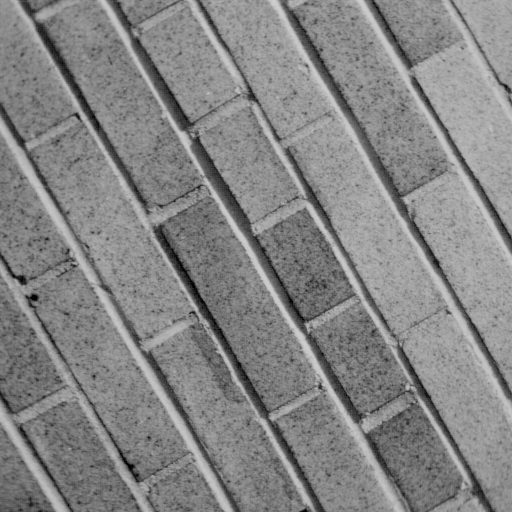}}
    \subfloat[90°]{
     \includegraphics[width=0.25\textwidth]{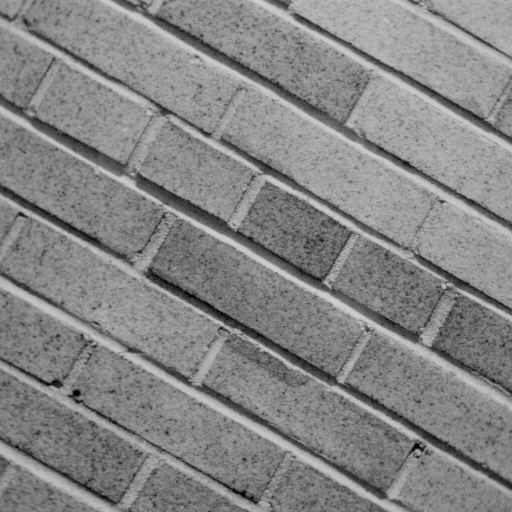}} \\
\subfloat[120°]{
     \includegraphics[width=0.25\textwidth]{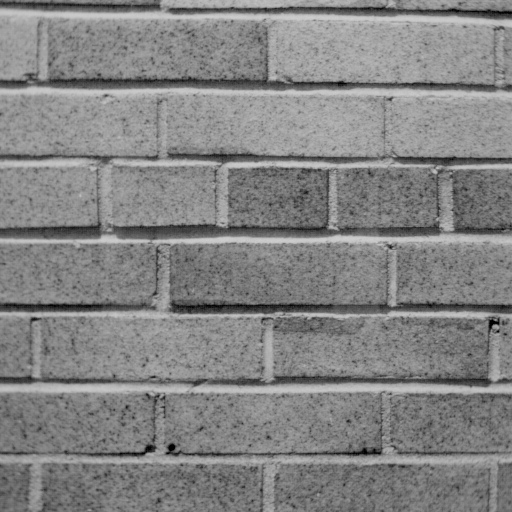}}
\subfloat[150°]{
    \includegraphics[width=0.25\textwidth]{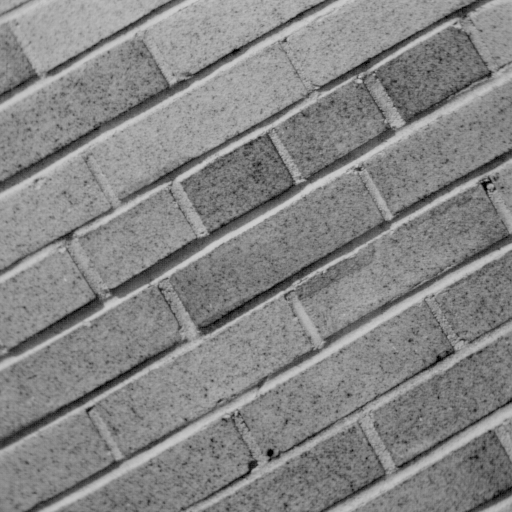}}    
\subfloat[200°]{
		\includegraphics[width=0.25\textwidth]{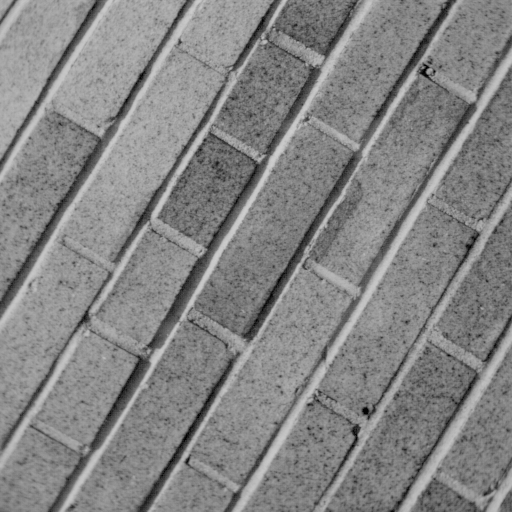}}
 \caption{A Normalized Brodatz texture and its rotated versions. Each panel is labeled with the degree of rotation.}
 \label{f:Rotate-Brodatz}
\end{figure}

It is worth to test the performance of the proposed method on textures rotated with other angles than only the multiples of
$\frac{\pi}{4}$.
The Signal and Image Processing Institute of the University of Southern California\footnote{Available at: \url{https://sipi.usc.edu/database/?volume=textures}} includes a set of images of rotated Brodatz textures by seven different angles: 0, 30, 60, 90, 120, 150 and 200 degrees. For this purpose, we selected 8 different textures with their respective rotations: Raffia, Straw, Weave, Wool, Leather, Water, Pigskin and Brick. This selection covers different kind of textures, making it representative of the overall Brodatz database.
Figure \ref{f:Rotate-Brodatz} shows an example of them (Brick).

%%%---------------------%%%

\begin{figure}
\hspace{-1.5cm}
    \includegraphics[scale=0.3]{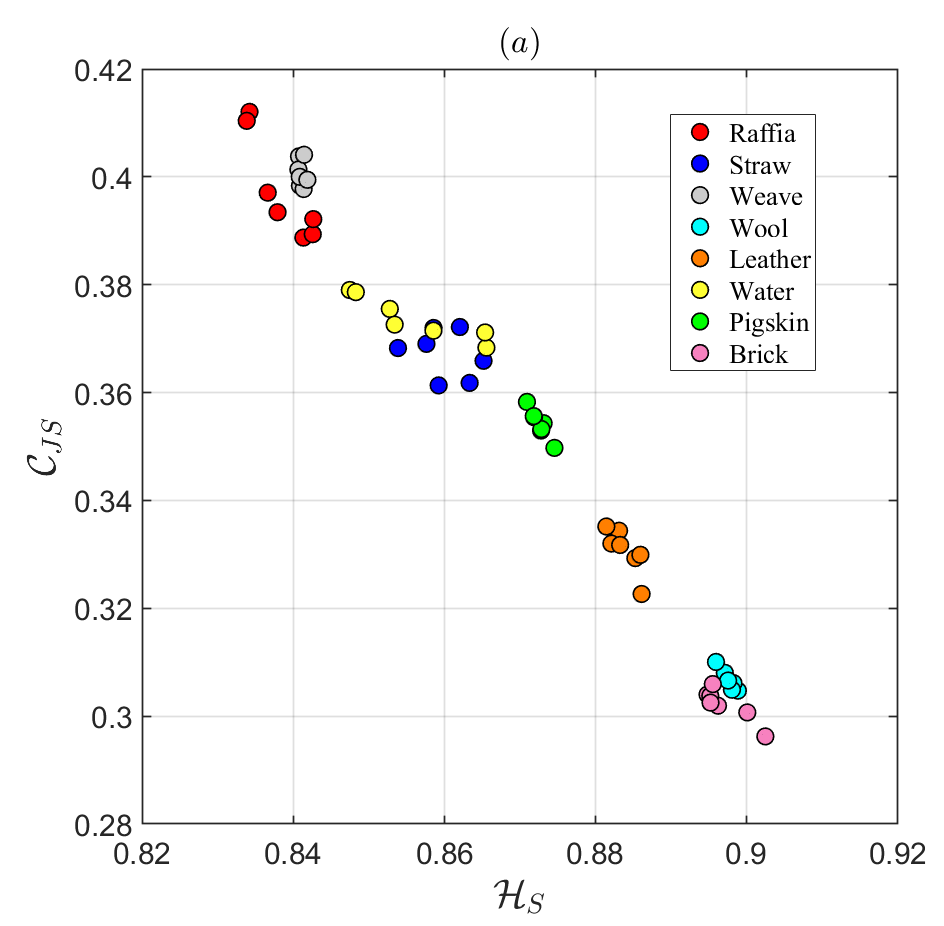}%
     \includegraphics[scale=0.3]{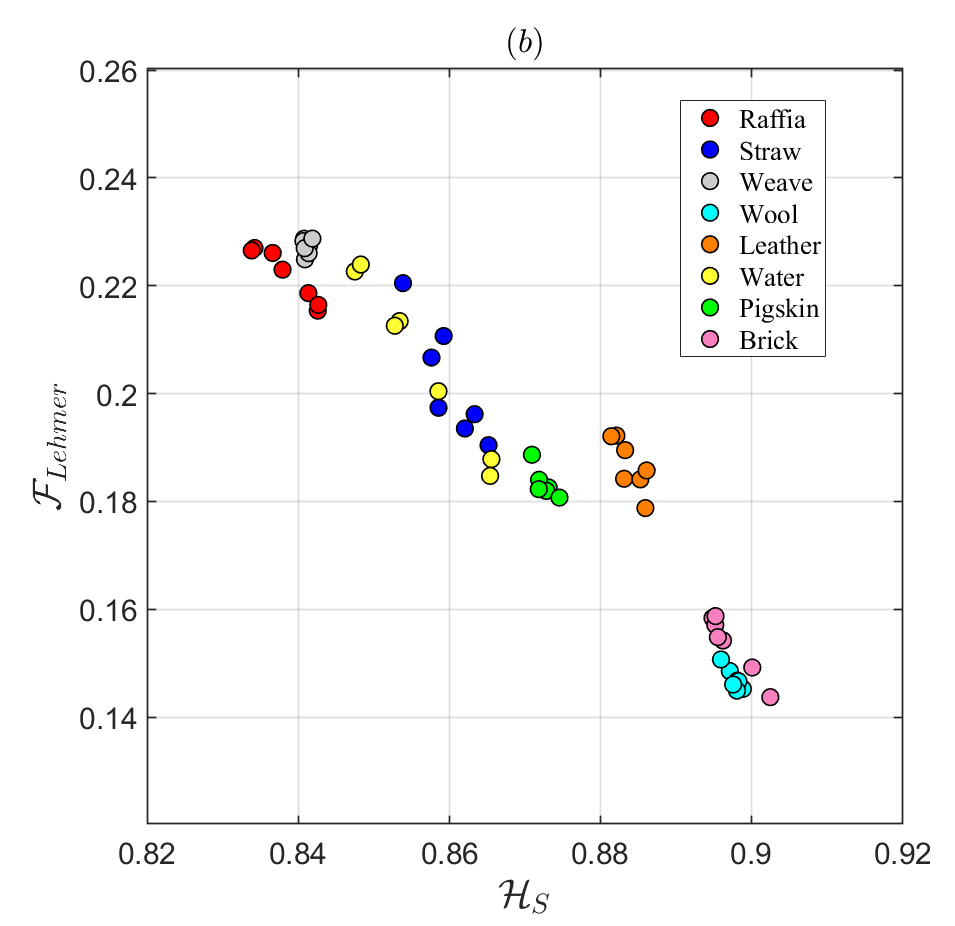}
    \caption{CECP (a) and FECP (b) for Rotated Brodatz set with $D\!=\!8$, $\tau\!=\!1$.}
    \label{f:Rotate-Brodatz-Entropy-Complexity}
\end{figure}

Figure \ref{f:Rotate-Brodatz-Entropy-Complexity} illustrates the CECP and the FECP representations for $D\!=\!8$ and $\tau\!=\!1$ of the 8 selected textures and their corresponding rotations. It can be observed that the locations in the both planes of each set of data points are fairly invariant to rotations, taking into account the loss of information. That is to say, the discrepancy that exists between some of the points within the same group is greater than in the previous examples. This is because the rotations by angles multiples of 90°, map squares into squares and so, ``all'' the information of the original image is necessarily preserved. In this case, an image is taken, it is rotated and then both itself and its rotated versions are cropped to a square of the same size. Thus, they don't share exactly the same information, although most of it.

%%%%%%%%%%%%%%%%%%%%%%%%%%%%%%%%%%%%%%%%%%%%%%%%%%%%%%%%
\section{Conclusions \label{sec:conclusions}}

Based on information quantifiers, we propose a new method to discern between different type of textures in images, which leverage the locality preserving properties of the Hilbert curve as a way to read them. We also add the calculation of the FIM quantifier and the FECP representation to existing literature as a complementary indicator to analyze complexity in two dimensional patterns. Its definition in Eq. \ref{Fisher-disc} depends on the $i$-order of a probability sequence $P\!=\!\{p_i\}$, and we show that this ambiguity can be overcome by choosing to scan the images
 using the Hilbert curve, whose advantage lies in that it avoids the directional biases found in conventional methods.

Briefly, the method is composed by three steps:
\begin{itemize}
\item Read the image using the Hilbert curve.
\item Compute the probability density function according to the Bandt \& Pompe symbolization method for embedding dimensions $5\!\le\!D\!\le\!8$ and delay $\tau\!=\!1$.
\item Compute the information theory quantifiers.
\end{itemize}
This method has been applied to simulated and real images with satisfactory results. 
Regarding the image of the SMM and its both ordered and randomized versions, the proposed method result efficient to distinguish between regular, aleatory and more complex designs. It results also a robust method against rotations and mirror-symmetry transformations, as shown in Fig. \ref{cecp-fisher-multif}.
Regarding the simulated $H$-fBs, we generated surfaces with different values of Hurst exponents, which occupied different locations in the CECP and FECP, and that also result invariant to rotation and mirror symmetry.

Regarding the real world images (Normalized and colored Brodatz data-bases), we observe a similar location in the planes, indicating that the proposed method is robust to the different color settings. Finally, we provide evidence that our discrimination method is also invariant to rotations. 

The invariance to rotation and color settings are important properties of an image classification method. Thus, the results presented in this paper will be useful for scientists working with diverse kinds of images and textures. 

Therefore, future works could include the analysis of medical images, aimed at classifying different pathologies. 
Another adventage of our method is its natural extension to higher dimensions by means of the $n$-dimensional Hilbert curve, as well as its possible adaptation to non-square images.

%%%%%%%%%%%%%%%%%%%%%%%%%%%%%%%%%%%%%%%%%%%%%%%%%%%%%%%%
\section*{Acknowledgements}
This work is partially supported by Programación Científica
UBACyT 2023-2024 (20020220200045BA and 20020220400162BA).
%%%%%%%%%%%%%%%%%%%%%%%%%%%%%%%%%%%%%%%%%%%%%%%%%%%%%%%%

%%%%%%%%%%%%%%%%%%%%%%%%%%%%%%%%%%%%%%%%%%%%%%%%%%%%%%%%


\begin{thebibliography}{}

\bibitem[Abdelmounaime and Dong-Chen, 2013]{AbdelmounaimeDong2013}
Abdelmounaime, S. and Dong-Chen, H. (2013).
\newblock New brodatz-based image databases for grayscale color and multiband
  texture analysis.
\newblock {\em International Scholarly Research Notices}, 2013(1):876386.

\bibitem[B.~Moon and Saltz, 2001]{Moon}
B.~Moon, H. V.~Jagadish, C.~F. and Saltz, J.~H. (2001).
\newblock Analysis of the clustering properties of the hilbert space-filling
  curve.
\newblock {\em IEEE Transactions on Knowledge and Data Engineering}, 13
  (1):124--141.

\bibitem[Bandt and Pompe, 2002]{BandtPompe02}
Bandt, C. and Pompe, B. (2002).
\newblock Permutation entropy: A natural complexity measure for time series.
\newblock {\em Physical Review Letters}, 88(17):174102.

\bibitem[Bariviera et~al., 2015]{Bariviera2015}
Bariviera, A.~F., Guercio, M.~B., Martinez, L.~B., and Rosso, O.~A. (2015).
\newblock A permutation information theory tour through different interest rate
  maturities: the libor case.
\newblock {\em Philosophical Transactions of the Royal Society of London.
  Series A, Containing Papers of a Mathematical or Physical Character},
  373:20150119--.

\bibitem[Brodatz, 1966]{Brodatz1966}
Brodatz, P. (1966).
\newblock {\em Textures : a photographic album for artists and designers}.
\newblock Dover publications, New York SE - XIV-114 p., plates, couv.
  illustrations [Don 15490] ; In-4° (27 cm).

\bibitem[Dai and Su, 2021]{Dai}
Dai, H. and Su, H. (2021).
\newblock Studies of norm-based locality measures of two-dimensional hilbert
  curves.
\newblock {\em SN Computer Science}, 2:403.

\bibitem[Evertsz and Mandelbrot, 1992]{Evertsz-Mandel}
Evertsz, C. and Mandelbrot, B. (1992).
\newblock {\em Multifractal Measures}, chapter Appendix B, in Chaos and
  Fractals-New Frontiers of Sciences, 1st. Ed.
\newblock Springer.

\bibitem[Falconer, 2014]{Falconer}
Falconer, K. (2014).
\newblock {\em Fractal Geometry - Mathematical Foundations and Applications,
  3rd. Ed.}
\newblock Wiley.

\bibitem[Feldman and Crutchfield, 1998]{FeldmanCrutchfield98}
Feldman, D.~P. and Crutchfield, J.~P. (1998).
\newblock Measures of statistical complexity: Why?
\newblock {\em Physics Letters A}, 238(4–5):244--252.

\bibitem[Feldman and Crutchfield, 2003]{Feldman2003}
Feldman, D.~P. and Crutchfield, J.~P. (2003).
\newblock Structural information in two-dimensional patterns: Entropy
  convergence and excess entropy.
\newblock {\em Physical Review E}, 67:051104.

\bibitem[Fernandes and Araújo, 2020]{FERNANDES2020109909}
Fernandes, L.~H. and Araújo, F.~H. (2020).
\newblock Taxonomy of commodities assets via complexity-entropy causality
  plane.
\newblock {\em Chaos, Solitons \& Fractals}, 137:109909.

\bibitem[Fisher, 1922]{Fisher1922}
Fisher, R.~A. (1922).
\newblock {On the Mathematical Foundations of Theoretical Statistics}.
\newblock {\em Philosophical Transactions of the Royal Society of London.
  Series A, Containing Papers of a Mathematical or Physical Character}, 222:pp.
  309--368.

\bibitem[Frieden, 1998]{Frieden1998}
Frieden, B.~R. (1998).
\newblock {\em Physics from {F}isher Information: A Unification}.
\newblock Cambridge University Press, Cambridge.

\bibitem[Hamouchene and Aouat, 2017]{Hamou-Aouat-2017}
Hamouchene, I. and Aouat, S. (2017).
\newblock Rotation-invariant method for texture matching using model-based
  histograms and {GLCM}.
\newblock {\em Int. J. Reasoning-based Intelligent Systems}, 9(1):3--11.

\bibitem[Keller and Sinn, 2005]{Keller2005}
Keller, K. and Sinn, M. (2005).
\newblock Ordinal analysis of time series.
\newblock {\em Physica A: Statistical Mechanics and its Applications},
  356:114--120.

\bibitem[Lamberti et~al., 2004]{Lamberti2004}
Lamberti, P., Martin, M., Plastino, A., and Rosso, O. (2004).
\newblock Intensive entropic non-triviality measure.
\newblock {\em Physica A: Statistical Mechanics and its Applications},
  334:119--131.

\bibitem[L\'opez-Ruiz et~al., 1995]{LMC95}
L\'opez-Ruiz, R., Mancini, H.~L., and Calbet, X. (1995).
\newblock A statistical measure of complexity.
\newblock {\em Physics Letters A}, 209(5–6):321--326.

\bibitem[Mandelbrot, 1982]{Mandelbrot}
Mandelbrot, B. (1982).
\newblock {\em The Fractal Geometry of Nature}.
\newblock W.H. Freeman and Company.

\bibitem[Mart\'{\i}n et~al., 2003]{Martin2003}
Mart\'{\i}n, M., Plastino, A., and Rosso, O. (2003).
\newblock Statistical complexity and disequilibrium.
\newblock {\em Physics Letters A}, 311(2–3):126 -- 132.

\bibitem[Micco et~al., 2008]{DeMicco2008}
Micco, L.~D., González, C., Larrondo, H., Martin, M., Plastino, A., and Rosso,
  O. (2008).
\newblock Randomizing nonlinear maps via symbolic dynamics.
\newblock {\em Physica A: Statistical Mechanics and its Applications},
  387:3373--3383.

\bibitem[Mischaikow et~al., 1999]{Mischaikow}
Mischaikow, K., Mrozek, M., Reiss, J., and Szymczak, A. (1999).
\newblock Construction of symbolic dynamics from experimental time series.
\newblock {\em Phys. Rev. Lett.}, 82:1144--1147.

\bibitem[Olivares et~al., 2012a]{OPR2012}
Olivares, F., Plastino, A., and Rosso, O.~A. (2012a).
\newblock Ambiguities in bandt–pompe’s methodology for local entropic
  quantifiers.
\newblock {\em Physica A}, 391(4):2518--2526.

\bibitem[Olivares et~al., 2012b]{Olivares2012B}
Olivares, F., Plastino, A., and Rosso, O.~A. (2012b).
\newblock Contrasting chaos with noise via local versus global information
  quantifiers.
\newblock {\em Physics Letters A}, 376:1577--1583.

\bibitem[Ribeiro et~al., 2012]{Ribeiro2012}
Ribeiro, H.~V., Zunino, L., Lenzi, E.~K., Santoro, P.~A., and Mendes, R.~S.
  (2012).
\newblock Complexity-entropy causality plane as a complexity measure for
  two-dimensional patterns.
\newblock {\em PLoS ONE}, 7(8):e40689.

\bibitem[Rosso et~al., 2007]{RossoNoise07}
Rosso, O.~A., Larrondo, H.~A., Mart\'{\i}n, M.~T., Plastino, A., and Fuentes,
  M.~A. (2007).
\newblock Distinguishing noise from chaos.
\newblock {\em Physical Review Letters}, 99(15):154102.

\bibitem[Rosso et~al., 2002]{Rosso2002}
Rosso, O.~A., Martin, M.~T., and Plastino, A. (2002).
\newblock Brain electrical activity analysis using wavelet-based informational
  tools.
\newblock {\em Physica A: Statistical Mechanics and its Applications},
  313:587--608.

\bibitem[Rosso et~al., 2013]{Rosso2013}
Rosso, O.~A., Olivares, F., Zunino, L., Micco, L.~D., Aquino, A.~L., Plastino,
  A., and Larrondo, H.~A. (2013).
\newblock Characterization of chaotic maps using the permutation bandt-pompe
  probability distribution.
\newblock {\em The European Physical Journal B 2013 86:4}, 86:1--13.

\bibitem[Saco et~al., 2010]{Saco2010}
Saco, P.~M., Carpi, L.~C., Figliola, A., Serrano, E., and Rosso, O.~A. (2010).
\newblock Entropy analysis of the dynamics of el niño/southern oscillation
  during the holocene.
\newblock {\em Physica A: Statistical Mechanics and its Applications},
  389:5022--5027.

\bibitem[Schwarz, 2024]{Lehmer}
Schwarz, K. (2024).
\newblock Factoradic permutation.
\newblock {\em
  http://www.keithschwarz.com/interesting/code/factoradic-permutation/FactoradicPermutation},
  Accessed 15/02/2024.

\bibitem[Shannon and Weaver, 1949]{book:shannon1949}
Shannon, E. and Weaver, W. (1949).
\newblock {\em The Mathematical Theory of Communication}.
\newblock University of Illinois Press, Champaign, IL.

\bibitem[Sigaki et~al., 2018]{Sigaki2018}
Sigaki, H.~Y., Perc, M., and Ribeiro, H.~V. (2018).
\newblock History of art paintings through the lens of entropy and complexity.
\newblock {\em PNAS}, 115(37):E8585--E8594.

\bibitem[Soriano et~al., 2011]{Soriano2011}
Soriano, M.~C., Zunino, L., Rosso, O.~A., Fischer, I., and Mirasso, C.~R.
  (2011).
\newblock Time scales of a chaotic semiconductor laser with optical feedback
  under the lens of a permutation information analysis.
\newblock {\em IEEE Journal of Quantum Electronics}, 47:252--261.

\bibitem[Sánchez-Moreno et~al., 2009]{Dehesa2009}
Sánchez-Moreno, P., Yanez, R.~J., and Dehesa, J.~S. (2009).
\newblock Discrete densities and fisher information.
\newblock In Bohner, M., Dosla, Z., Ladas, G., Ünal, M., and Zafer, A.,
  editors, {\em Proceedings of the 14th International Conference on Difference
  Equations and Applications}, pages 291--298. Bahcesehir University Press.

\bibitem[Tang et~al., 2024]{Tang}
Tang, C., Chen, J., and Wang, J. (2024).
\newblock Image encryption algorithm based on 2d-linear-infinite-collapse
  chaotic map and improved hilbert curve.
\newblock {\em Internationa Journal of Bifurcation and Chaos}, 34(6):2450067.

\bibitem[Vignat and Bercher, 2003]{Vignat2003}
Vignat, C. and Bercher, J.-F. (2003).
\newblock Analysis of signals in the fisher–shannon information plane.
\newblock {\em Physics Letters A}, 312:27--33.

\bibitem[Zunino and Ribeiro, 2016]{ZUNINO2016679}
Zunino, L. and Ribeiro, H.~V. (2016).
\newblock Discriminating image textures with the multiscale two-dimensional
  complexity-entropy causality plane.
\newblock {\em Chaos, Solitons \& Fractals}, 91:679--688.

\bibitem[Zunino et~al., 2010]{ZUNINO2010efficiency}
Zunino, L., Zanin, M., Tabak, B.~M., Pérez, D.~G., and Rosso, O.~A. (2010).
\newblock Complexity-entropy causality plane: A useful approach to quantify the
  stock market inefficiency.
\newblock {\em Physica A: Statistical Mechanics and its Applications},
  389(9):1891--1901.

\end{thebibliography}
\end{document}